%% file: iclr2026_conference.tex
\documentclass{article} % For LaTeX2e
\usepackage{iclr2026_conference,times}

\input{math_commands.tex}

\usepackage{hyperref}
\usepackage{url}
\usepackage{graphicx}
\usepackage{booktabs}
\usepackage{multirow}
\usepackage{xcolor}
\usepackage{colortbl} % Required for \rowcolor
\usepackage[table]{xcolor} % For coloring rows
\usepackage{multicol} % Enable multi-column environment
\usepackage{tabulary}
\usepackage{subcaption}
\usepackage{amsmath} % Recommended for mathematical typesetting
\usepackage{amssymb} % Required for \lesssim
\usepackage[noend]{algpseudocode} % noend option for cleaner look
\usepackage{algorithm}
\usepackage{tabularx}   % For creating tables with auto-wrapping text
\DeclareCaptionSubType{algorithm} % enables subalgorithm
\usepackage[most]{tcolorbox} % Load the tcolorbox package

\usepackage[utf8]{inputenc} % allow utf-8 input
\usepackage[T1]{fontenc}    % use 8-bit T1 fonts
\usepackage{amsfonts}       % blackboard math symbols
\usepackage{nicefrac}       % compact symbols for 1/2, etc.
\usepackage{microtype}      % microtypography
\usepackage{bm}
\usepackage{amsthm}

\usepackage{algorithmicx}
\usepackage{algpseudocode}

\usepackage{pifont}
\usepackage{diagbox}
\usepackage{wrapfig}

\usepackage{fontawesome5}    % 支持 \faFire 等图标
\usepackage{pifont}          % 如 \textcircled 可选增强
\definecolor{custompurple}{RGB}{223,213,230}
\definecolor{customorange}{RGB}{251,231,207}
\definecolor{softgray}{RGB}{245,245,245}     % 极浅灰
\definecolor{midgray}{RGB}{220,220,220}      % 中灰
\definecolor{darkgray}{RGB}{180,180,180}     % 深灰，保留层次感
\definecolor{softyellow}{RGB}{255,249,196}   % 柔和浅黄（像纸张）
\definecolor{lemonyellow}{RGB}{253,233,124}  % 柠檬黄
\definecolor{mintgreen}{RGB}{212,237,218}    % 薄荷绿（柔和）
\definecolor{springgreen}{RGB}{120,200,150}  % 科研绿（常用于模型优于 baseline）
\definecolor{salmonred}{RGB}{255,204,204}    % 柔和红（用于背景）
\definecolor{sciencecrimson}{RGB}{220,50,47} % 深红（用于字体或线条）
\definecolor{lavender}{RGB}{230,230,250}     % 薰衣草紫（浅、优雅）
\definecolor{sciencepurple}{RGB}{178,102,255} % 科研常用紫

\definecolor{apricotorange}{RGB}{255,223,186}   % 杏橘色（浅橘）
\definecolor{deeporange}{RGB}{255,140,0}        % 深橘（对比强烈）

\newcommand{\cmark}{\ding{51}} % ✓
\usepackage{diagbox}
\usepackage{arydshln} 
\usepackage{tikz}

\newtheorem{definition}{Definition}[section]

\renewcommand{\thefootnote}{\fnsymbol{footnote}}

\usepackage{comment}

\title{
Action-aware Dynamic Pruning for Efficient Vision-Language-Action
Manipulation
}

\renewcommand{\thefootnote}{\fnsymbol{footnote}}

\author{%
  Xiaohuan Pei$^{1*}$\quad
  Yuxing Chen$^{1*}$\quad
  Siyu Xu$^{1}$\quad
  Yunke Wang$^{1}$\quad
  Yuheng Shi$^{1}$\quad
  Chang Xu$^{1}$
  \\
  $^{1}$School of Computer Science, The University of Sydney\\
  \texttt{\{xiaohuan.pei, s.xu, yunke.wang, c.xu\}@sydney.edu.au}\\
  \texttt{\{yche0009, yshi0087\}@uni.sydney.edu.au}
}

\iclrfinalcopy % Uncomment for camera-ready version, but NOT for submission.
\begin{document}

\maketitle
% （可选）从这里开始把后续脚注恢复成阿拉伯数字
\setcounter{footnote}{0}
\renewcommand{\thefootnote}{\arabic{footnote}}

\input{sec/0_abstract}    
\input{sec/1_intro}

\input{sec/2_related}
\input{sec/3_method}

\input{sec/4_experiment}

\input{sec/5_conclusion}
\bibliography{iclr2026_conference}
\bibliographystyle{iclr2026_conference}
\clearpage
\appendix
\input{sec/x_supp}

\end{document}

%% file: math_commands.tex
\usepackage{amsmath,amsfonts,bm}

\def\eqref#1{equation~\ref{#1}}
\def\1{\bm{1}}

\DeclareMathAlphabet{\mathsfit}{\encodingdefault}{\sfdefault}{m}{sl}
\SetMathAlphabet{\mathsfit}{bold}{\encodingdefault}{\sfdefault}{bx}{n}

%% file: sec/0_abstract.tex
\begin{abstract}
Robotic manipulation with Vision-Language-Action models requires efficient inference over long-horizon multi-modal context, where attention to dense visual tokens dominates computational cost. Existing methods optimize inference speed by reducing visual redundancy within VLA models, but they overlook the varying redundancy across robotic manipulation stages. We observe that the visual token redundancy is higher in coarse manipulation phase than in fine-grained operations, and is strongly correlated with the action dynamic. 
Motivated by this observation, we propose \textbf{A}ction-aware \textbf{D}ynamic \textbf{P}runing (\textbf{ADP}), a multi-modal pruning framework that integrates text-driven token selection with action-aware trajectory gating. Our method introduces a gating mechanism that conditions the pruning signal on recent action trajectories, using past motion windows to adaptively adjust token retention ratios in accordance with dynamics, thereby balancing computational efficiency and perceptual precision across different manipulation stages. 
Extensive experiments on the LIBERO suites and diverse real-world scenarios demonstrate that our method significantly reduces FLOPs and action inference latency (\textit{e.g.} 1.35× speed up on OpenVLA-OFT) while maintaining competitive success rates (\textit{e.g.} 25.8\% improvements with OpenVLA) compared to baselines, thereby providing a simple plug-in path to efficient robot policies that advances the efficiency and performance frontier of robotic manipulation. Our project website is: \href{https://vla-adp.github.io/}{ADP.com}.
\end{abstract}

%% file: sec/1_intro.tex
\section{Introduction}
\label{sec:Introduction}

Large vision language models~\cite{liu2023llava, liu2023improvedllava, liu2024llavanext,team2023gemini,awadalla2023openflamingo} have recently been extended into \textbf{V}ision–\textbf{L}anguage-\textbf{A}ction (VLA) models~\cite{kim2024openvla, kim2025fine,black2024pi,li2024cogact,brohan2024rt,wen2025diffusionvla,wen2025dexvla,bjorck2025gr00t} that map both the visual observation and language instruction to executable robot actions. In the mainstream pipeline, a vision encoder produces dense visual tokens from one or more camera views, a projector aligns them to the language space, and an LLM fuses all modalities to predict actions. However, this multi-modal design introduces long input sequences with numerous visual tokens that are only weakly relevant to the current manipulation operation, 
which inflates compute, memory footprint, and latency, and it can dilute attention over truly task-relevant cues.

\begin{figure*}[t]
\centering
% \vspace{-0.1in}
\hspace*{-0.5cm}
\includegraphics[width=0.99\linewidth]{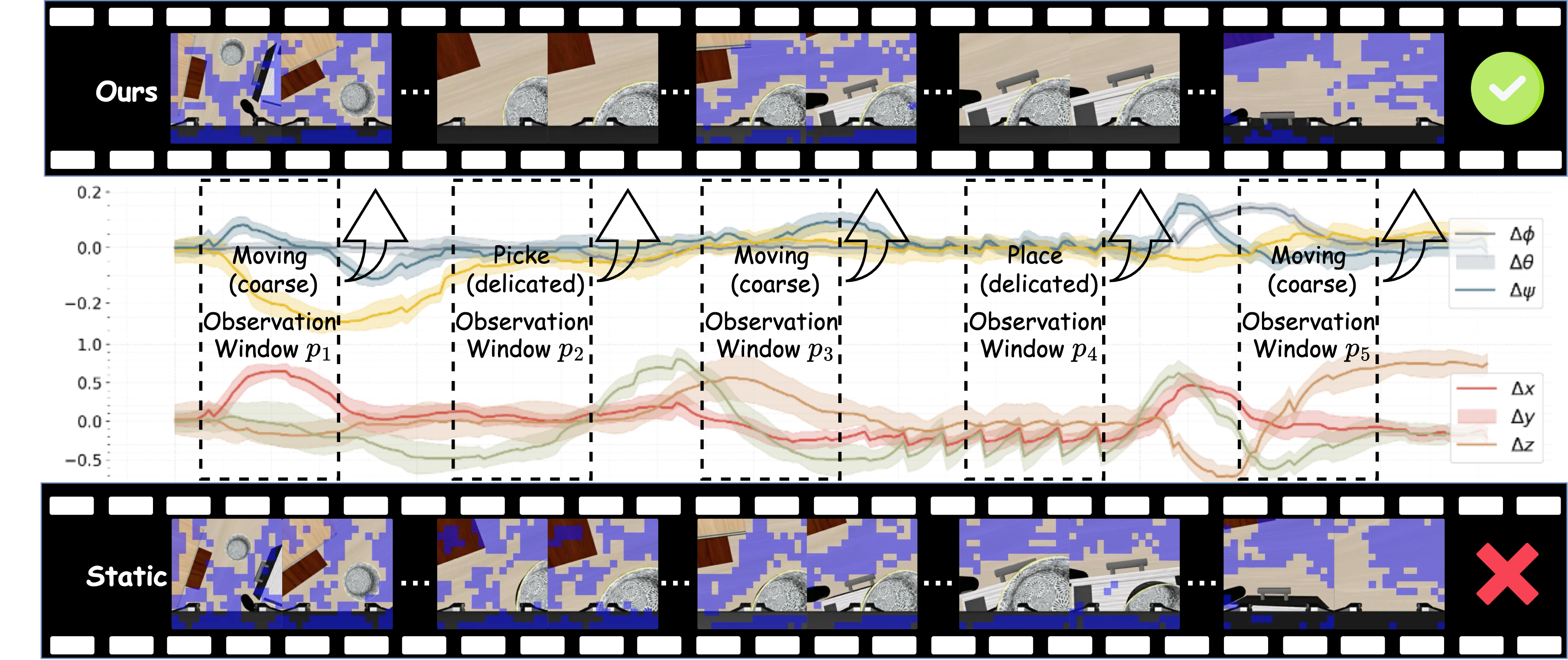} 
\caption{\textbf{Action-aware dynamic pruning vs.\ static pruning.} We visualize five \emph{past} observation windows as cases from a manipulation episode that condition the \emph{current anticipatory} window. $p_1$, $p_3$ and $p_5$ reflect coarse phases, prompting the gate to \emph{enable} pruning to suppress redundant tokens, whereas $p_2$ and $p_4$ are delicate phases requiring detail vision context, so pruning is \emph{disabled} and full vision is used. 
The curves depict robot's motion that drives the gating rule.}
\label{fig:motivation}
\vspace{-0.15in}
\end{figure*}

Existing work pursues efficiency via architectural lightening and modality-aware compression, such as RoboMamba \cite{liu2024robomamba} that focuses on lightweight designs, DeeR-VLA \cite{yue2024deer} that aims at structured pruning/reparameterization, Mole-VLA \cite{zhang2025mole}  that targets conditional layer activation, VLA-Cache \cite{xu2025vla} that focuses on cache reuse, and EfficientVLA \cite{yang2025efficientvla} that aims to prune visual tokens via attention.
% Existing work ...
However, a key but underexplored property of robotic manipulation is that \textbf{visual redundancy in VLAs is action-aware across different manipulation stages}. 
As Fig.~\ref{fig:motivation} shows, during coarse-grained operations (\textit{e.g.}, relocating), global movement dominates and redundant tokens can be pruned; during fine-grained phases (\textit{e.g.}, grasping), local geometry and detailed cues dominate and preserving full vision is preferred.
Moreover, the relevance of visual patches is not only text conditioned (semantics of the instruction) but also \emph{action conditioned} (instantaneous end-effector motion and gripper state). Treating all steps uniformly, or ranking tokens solely by mixed attention scores, therefore yields suboptimal pruning schedules that either prune too little (limited savings) or prune too much (accuracy loss), especially in multi-view settings (scene and wrist/gripper cameras) where importance is unevenly distributed across 
% views and 
time.

% As Fig. \ref{fig:motivation} shown, 
% during coarse motion (\textit{e.g.}, reaching or relocating), many visual tokens convey overlapping context and can be safely downselected; during fine-grained operations (\textit{e.g.}, grasp, align, insert), small details such as contacts, edges, apertures become critical, and aggressive pruning risks failure. 

% As Fig.~\ref{fig:motivation}, during coarse-grained operations (\textit{e.g.}, grasp in some tasks), local geometry dominates and surrounding redundant tokens can be safely downselected; during fine-grained motion (\textit{e.g.}, relocating in some cases), maintaining full vision to preserve global spatial context is preferred. 

% As Fig.~\ref{fig:motivation} shows, during coarse-grained operations (\textit{e.g.}, relocating in some cases), global movement dominates and many surrounding tokens become redundant, so pruning can be \emph{enabled} to suppress them; during fine-grained phases \textit{e.g.}, grasp in some tasks), local geometry and detailed visual cues are critical, so pruning is \emph{disabled} and full vision is preserved. 

% As Fig.~\ref{fig:motivation} shows, during coarse-grained operations (\textit{e.g.}, relocating), global movement dominates and many surrounding tokens are redundant, so pruning is \emph{enabled}; in fine-grained phases (\textit{e.g.}, grasping), local geometry and detailed cues are crucial, so pruning is \emph{disabled} to preserve full vision.

% \textbf{Action-aware dynamic pruning vs.\ static pruning.} We visualize three \emph{past} observation windows as cases from a manipulation episode that condition the \emph{current anticipatory} window. $p_1$ and $p_3$ reflect coarse local phases 
% % (pick/place)
% , prompting the gate to \emph{enable} pruning to suppress redundant tokens, whereas $p_2$ is a delicate moving phase requiring global spatial context, so pruning is \emph{disabled} and full vision is used. 
% The curves depict robot's motion that drives the gating rule. 

% and the thumbnails illustrate the tokens retained under each setting; the static baseline applies a fixed mask throughout.

% \caption{
% Comparison of our action-aware dynamic method \textit{vs.} previous static accelerating method. We shows three observation windows cases for an robotic manipulation episode. $p_1$ and $p_3$ correspond to delicate local operations (pick/place), where pruning is \emph{enabled} to suppress redundant tokens; $p_2$ is a coarse moving phase that requires global spatial context, so pruning is \emph{disabled}. Curves indicate EEF motion that drive the gate, and thumbnails illustrate the tokens retained.
% }

To address this challenge, we introduce Action-aware Dynamic Pruning (ADP), a plug-and-play strategy that reduces computation while preserving manipulation fidelity. ADP is built on two complementary ideas: (1) \emph{Text-driven Pruning} evaluates the relevance of visual patch using cross-modal similarities, selecting only the most relevant tokens before entering deep fusion in the subsequent layers. (2) \emph{Action-aware Dynamics} modulate whether pruning is activated at a given step using a lightweight decision signal derived from the end-effector trajectory within each action window. 
Specifically, when the recent motion magnitude is relatively low compared to past motion statistics (delicate phases), pruning is disabled to preserve the full visual field for precise control. Conversely, when the motion magnitude is relatively high compared to past motion statistics (coarse phases), pruning is engaged to suppress redundancy and save FLOPs.
We implement a gated mechanism that treats recent action statistics as a pruning signal over sliding trajectory windows, adaptively adjusting retention ratios according to motion dynamics and balancing efficiency with precision across manipulation stages. Our contributions can be summarized as:
(1) We show that the importance of the visual token in VLA models varies within different stages of robotic manipulation. This insight motivates our dynamic pruning method tailored to manipulation phases compared to static pruning approaches.
(2) We propose \emph{text-driven action-aware pruning} that combines task instruction relevance with a gating rule based on end-effector motion, enabling adaptive switching between pruned and full-vision states.
(3) We present a principled complexity analysis and extensive experiments in simulation and real-world settings, demonstrating that our method reduces FLOPs and latency while maintaining fine visual details required for successful manipulation.

\section{Preliminary}
\label{sec:Preliminary}
The mainstream vision–language–action paradigm extends large vision–language models to generate executable robot actions from multi-modal inputs—visual observations (scene and wrist/gripper views) and task instructions. A pre-trained vision encoder produces visual tokens, a projector aligns them to the LLM token space, and the LLM fuses modalities and autoregressively emits 
action tokens, which 
are de-tokenized into a continuous 7-dimensional robot action.

Formally, given a sample of a scene image $I^s \in \mathbb{R}^{H \times W \times 3}$, a gripper view $I^g \in \mathbb{R}^{H \times W \times 3}$, and a task instruction $I^t \in \mathbb{R}^{N \times L}$, the vision encoder $f^v_{\text{enc}}$ (DINOv2~\cite{oquab2023dinov2} and SigLIP~\cite{zhai2023sigmoid}) and text tokenizer $f^t_{\text{enc}}$ project the multi-modal data into the same latent dimension space:
\begin{equation}
\label{eq:projecter}
\mathbf{X}^{\text{vis}} = f^v_{\text{enc}}(I^s, I^g), \quad \mathbf{X}^{\text{txt}} = f^t_{\text{enc}}(I^t),
\end{equation}
where $\mathbf{X}^{\text{vis}} \in \mathbb{R}^{L_{\text{vis}} \times D}$ and $\mathbf{X}^{\text{txt}} \in \mathbb{R}^{L_{\text{txt}} \times D}$ represent latent embeddings for vision and text.

\textbf{OpenVLA.} In the latent space, the embedding representations are concatenated into a multi-modal sequence as:
\begin{equation}
\label{eq:pre:emb}
\mathbf{X}^m = \mathbf{X}^{[BOS]} \oplus \mathbf{X}^{\text{vis}} \oplus \mathbf{X}^{\text{prop}} \oplus \mathbf{X}^{\text{txt}} 
\end{equation}
where $\oplus$ denotes concatenating embeddings of [BOS], vision, text, and proprioceptive (optionally) along the sequence length dimension, yielding multi-modal inputs $\mathbf{X}^m \in \mathbb{R}^{1+L_{\text{vis}}+L_{\text{txt}}+1}$. 
The multi-modal sequence $\mathbf{X}^m$ is then fed into a Large Language Model (LLM) $f_{\text{LLM}}$ (Llama2~\cite{touvron2023llama}), which performs contextual reasoning and autoregressively generates an action token sequence:
\begin{equation}
\label{eq:llm}
    p(\hat{\mathbf{a}} \mid \mathbf{X}^m) 
    = \prod_{j=1}^{7} f_{\text{LLM}}(\hat a_j \mid \mathbf{X}^m, \hat a_{<j}),
\end{equation}
where $\hat{\mathbf{a}} = (\hat a_1,\ldots,\hat a_7) \in \mathbb{T}^7$ and $\mathbb{T}=\{1,\ldots,K\}$ denotes the reserved action token space.

\textbf{OpenVLA-OFT.}
For optimize finetuning version, the paradigm shifts to parallel decoding without traditional autoregressive decoding stage. In the latent space, the method introduces $L_{\text{act}}$ placeholder $\mathbf{X}^{\text{place}}$ as inputs for actions positions:
\begin{equation}
    \label{eq:pre:emb0}
    \mathbf{X}^m = \mathbf{X}^{[\text{BOS}]}  \oplus \mathbf{X}^{\text{vis}} \oplus \mathbf{X}^{\text{prop}} \oplus \mathbf{X}^{\text{txt}}  \oplus \mathbf{X}^{\text{place}},
\end{equation}
where $\mathbf{X}^m \in \mathbb{R}^{1+L_{\text{vis}}+L_{\text{txt}}+1+L_{\text{act}}}$ and the first placeholder as current action placeholder and another $L_{\text{act}}-1$ actions as future actions.
During generation, the LLM directly predicts action tokens at action chunk in parallel:
\begin{equation}
\label{eq:oft_parallel}
    \hat{\mathbf{A}} = f_{\text{LLM}}(\mathbf{X}^m)\big|_{\text{place}} \in \mathbb{T}^{L_{\text{act}} \times 7},
\end{equation}
where $\hat{\mathbf{A}} = [\hat{\mathbf{a}}_1;\ldots;\hat{\mathbf{a}}_{L_{\text{act}}}]$, and each $\hat{\mathbf{a}}_i = (\hat a_{i,1},\ldots,\hat a_{i,7}) \in \mathbb{T}^7$ represents the tokenized $7$-DoF action predicted at the $i$-th placeholder position.  To expand action representation ability, the oft version also introduce the 
% To connect
continuous control with token-level modeling, which each dimension of the continuous action
is uniformly discretized into 256 bins. 
During generation, the 7 degrees of freedom action follows the standard 
gripper parameterization:
\begin{equation}
    \hat{\mathbf{a}}^{\,c} = [\,\Delta x,\ \Delta y,\ \Delta z,\ \Delta\phi,\ \Delta\theta,\ \Delta\psi,\ g\,]^\top,
\end{equation}
where $(\Delta x,\Delta y,\Delta z)$ denote Cartesian displacements, $(\Delta\phi,\Delta\theta,\Delta\psi)$ denote Euler-angle rotations (roll, pitch, yaw), and $g$ denotes the gripper.

%% file: sec/2_related.tex
\section{Related Work}
\label{sec:Related Work}
\textbf{Vision-Language-Actions.}
Vision-Language-Action (VLA) models extend large vision-language models with an action generator that links multimodal perception to low-level robot control. They take images and text instructions, encode them into a shared latent space, then decode it into executable actions.
Early approaches~\cite{chi2023diffusion,kim2024openvla} map multimodal embeddings to discrete tokens, while more recent works~\cite{wen2025diffusionvla,kim2025fine} emphasize continuous parallel decoding actions for improving performance. To this end, lightweight MLPs, diffusion-based decoders, and parallel decoding strategies have been introduced to transform hidden states into temporally consistent trajectories. Representative architecture include CogACT \cite{li2024cogact}, OpenVLA \cite{kim2024openvla}, OpenVLA-OFT \cite{kim2025fine},  
and $\pi$ series \cite{black2024pi}, 
which adopt diffusion modules for iterative refinement of continuous actions, as well as optimized fine-tuning frameworks that leverage placeholder tokens for parallel action prediction.

\textbf{Efficient Robotic Manipulations.} 
The high computational complexity of Vision-Language-Action (VLA) models poses significant efficiency challenges for real-time robotic control. 
To address this, recent works have proposed efficiency-oriented strategies that can be broadly divided into training-aware and training-free approaches. 
Training-aware methods, such as RoboMamba \cite{liu2024robomamba} and DeeR-VLA \cite{yue2024deer}, redesign architectures or apply compression and pruning during training, achieving notable speedups while preserving accuracy. For instance, DeeR-VLA \cite{yue2024deer} introduces dynamic reparameterization and structured pruning to reduce FLOPs, and Mole-VLA \cite{zhang2025mole} selectively activates subsets of model layers conditioned on task demands, yielding scalable deployment. On the other hand, training-free methods aim to accelerate inference without retraining. VLA-Cache \cite{xu2025vla} reuses keys and values of uninformative tokens between consecutive steps to save computation, while EfficientVLA \cite{yang2025efficientvla} retain task-relevant patches identified via attention maps. 
However, these methods typically rely on single-layer heuristics or static rules that may overlook stage-dependent redundancy in manipulation tasks. 

%% file: sec/3_method.tex
\section{Methodology}
\label{sec:Methodology}
\begin{figure*}[t]
\centering
\includegraphics[width=0.99\linewidth]{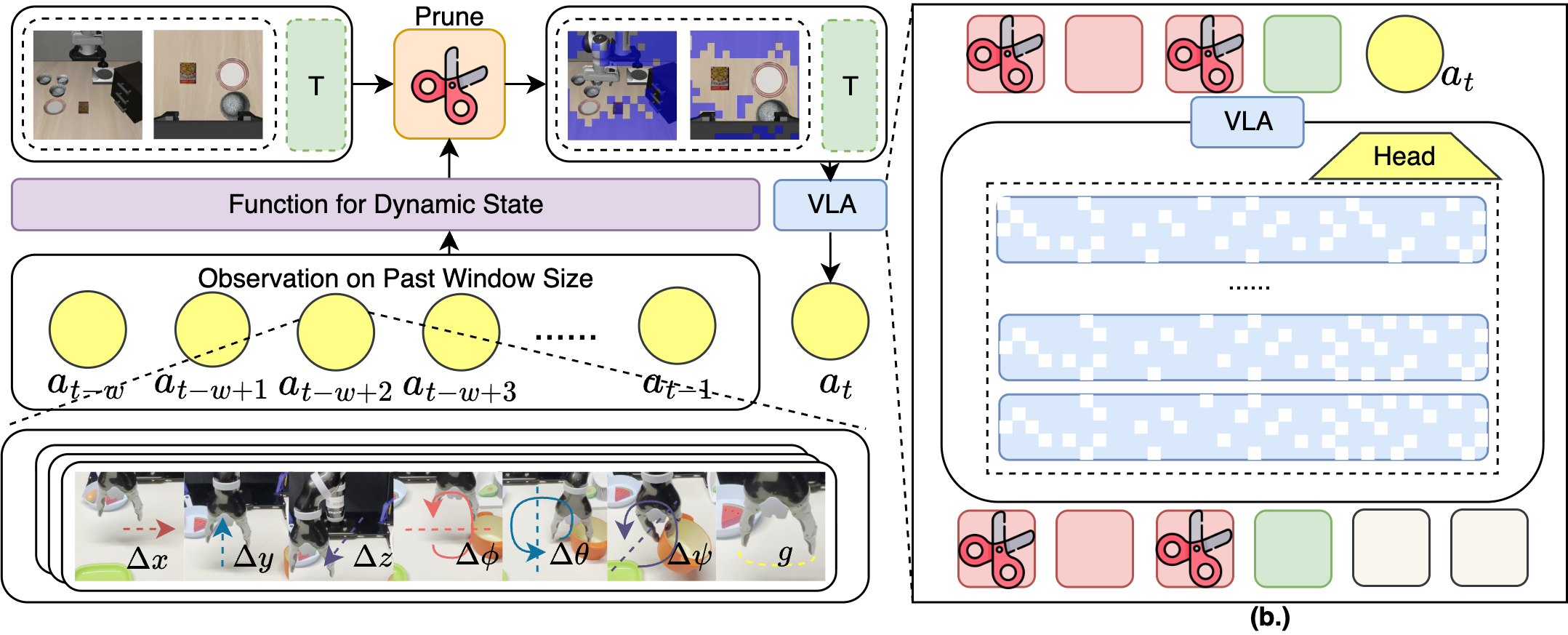} 
\caption{
Overview of our proposed \textbf{A}ction-aware \textbf{D}ynamic \textbf{P}runing (\textbf{ADP}) for Vision-Language-Action models.
\textbf{(a.)} Action-aware gating: the pruning function adaptively determines whether to prune based on recent end-effector trajectories 
($\Delta x, \Delta y, \Delta z, \Delta \phi, \Delta \theta, \Delta \psi, g$) 
, enabling dynamic pruning. \textbf{(b.)} Anticipatory pruning: task-relevant visual tokens are selected via attention-based relevance, while redundant patches are discarded before entering the VLA backbone. 
}
\label{fig:method}
\vspace{-0.1in}
\end{figure*} 

In this section, we introduce our proposed method Action-aware Dynamic Pruning (ADP) in VLAs for robotic manipulations.
We first introduce text-driven pruning that identifies text-relevant tokens in Section \ref{subsec:Text-driven Anticipatory Pruning} and then introduce action-aware dynamics that dynamically modulates the pruning strategy according to the past observed actions in Section \ref{secsec:Dynamic}.

\subsection{Text-driven Anticipatory Pruning}
\label{subsec:Text-driven Anticipatory Pruning}
\begin{wrapfigure}{r}{0.45\textwidth}
  \vspace{-0.1in}
  \setlength{\abovecaptionskip}{0.00cm}
  \setlength{\belowcaptionskip}{0.00cm}
  \begin{center}
    \includegraphics[width=0.44\textwidth]{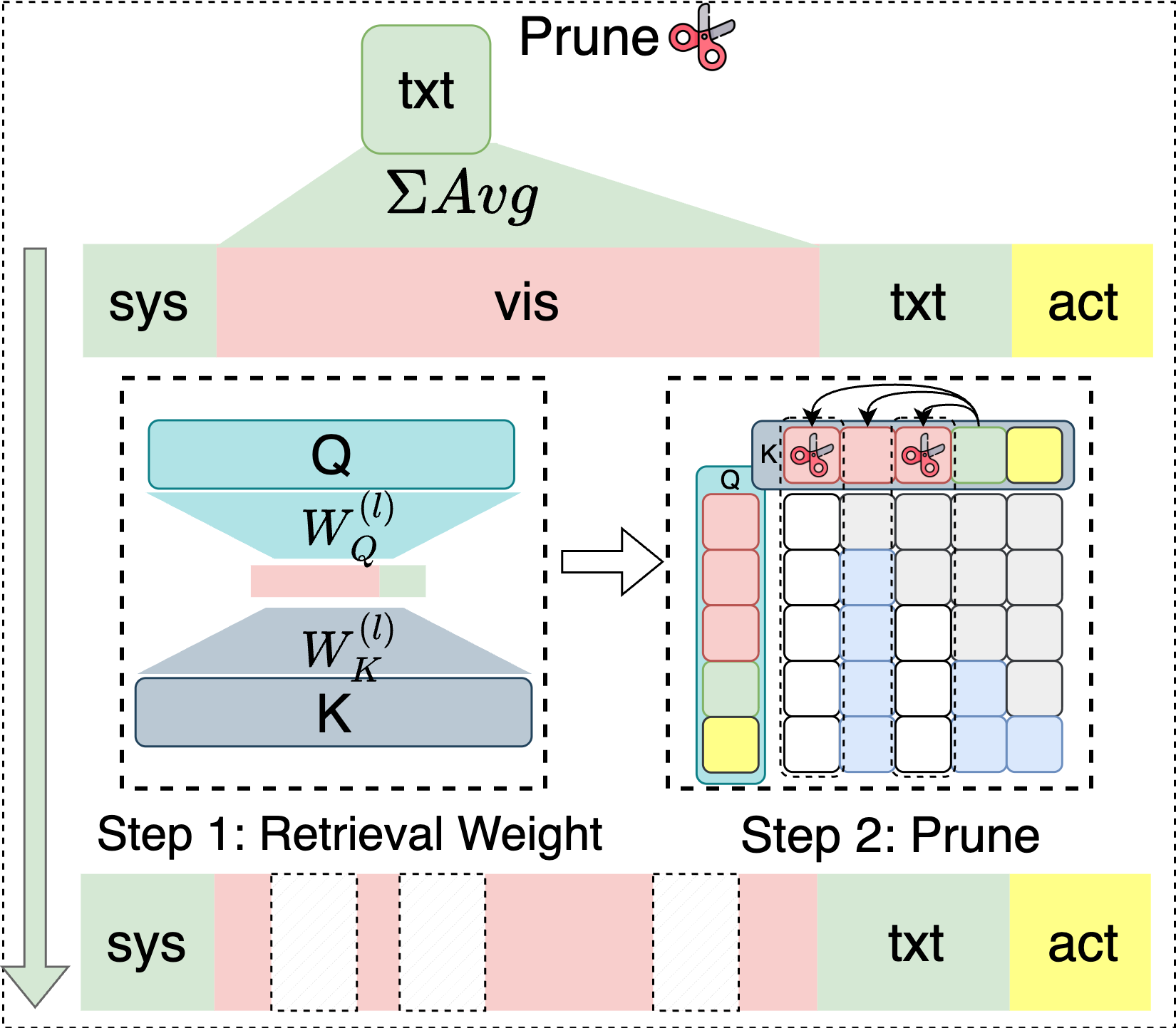}
  \end{center}
    \caption{Text-driven Anticipatory Pruning.
    \textbf{Step 1}: Retrieval pretrained 
    % $W^{(l)}_{Q}$ / $W^{(l)}_{K}$
    weights
    from Layer $l$ 
    to compute relevance scores.
    \textbf{Step 2}: Treat text as a guider to prune vision tokens
    based on the ranking
    .
    }
    \label{fig:prune}
  \vspace{-0.55in}
\end{wrapfigure}

Before entering the LLM, the multimodal sequence $\mathbf{X}$ still contains a large number of redundant visual tokens, which increase computation and dilute attention focus. 
To mitigate this redundancy, we compute the relevance of visual tokens with respect to task instructions at each layer, which shown in Figure~\ref{fig:prune}. Let the hidden state at layer $l$ be $\mathbf{H}^{(l)} \in \mathbb{R}^{S \times D}$ ($\mathbf{H}^{(0)}=\mathbf{X}^{m}$ in Eq. \ref{eq:pre:emb}). We partition $\mathbf{H}^{(l)}$ into the vision and text subsets $\mathbf{H}^{(l)}_{\text{vis}} \in \mathbb{R}^{L^{\text{vis}} \times D}$ and $\mathbf{H}^{(l)}_{\text{txt}} \in \mathbb{R}^{L^{\text{txt}} \times D}$. Applying the projection matrices, we obtain query and key representations:
\begin{equation}
\mathbf{Q}^{(l)} = \mathbf{H}^{(l)}_{\text{txt}} W_Q^{(l)}, \quad
\mathbf{K}^{(l)} = \mathbf{H}^{(l)}_{\text{vis}} W_K^{(l)},
\end{equation}
reshaped into multi-head form $\mathbf{Q}^{(l)} \in \mathbb{R}^{N^h \times L^{\text{txt}} \times d}$ and $\mathbf{K}^{(l)} \in \mathbb{R}^{N^h \times L^{\text{vis}} \times d}$. The scaled dot-product similarity is then computed as
\begin{equation}
\mathbf{A}^{(l)} = \frac{\mathbf{Q}^{(l)} (\mathbf{K}^{(l)})^\top}{\sqrt{d}}
\in \mathbb{R}^{N^h \times L^{\text{txt}} \times L^{\text{vis}}},
\end{equation}
where each entry measures the degree to which a text token attends to a visual patch. 
To derive a global importance score per visual token, we average across heads and text queries:
\begin{equation}
\label{eq:attention score}
\Phi^{(l)}(v) 
= \frac{1}{N^h \cdot L_{\text{txt}}} 
\sum_{h=1}^{N^h} \sum_{t=1}^{L_{\text{txt}}} \mathbf{A}^{(l)}_{h,t,v}, 
\end{equation}
\begin{equation}
\label{eq:keep embeddings}
\mathbf{X}_{\text{keep}} = \operatorname{Top-K}\big(\Phi^{(l)}, k\big), \quad k = \lfloor \rho \cdot L^{\text{vis}} \rfloor, 
\end{equation}
where $v \in \{1, \dots, L_{\text{vis}}\}$ indexes visual tokens,  $N^h$ is the number of attention heads, $L^{\text{txt}}$ is the text sequence length,  
and $L^{\text{vis}}$ is the total number of visual tokens.
In multi-view scenarios with $C$ input images (\textit{e.g.}, scene and wrist views), 
each contributing $L^{\text{vis}}_c$ patches such that
$\sum_{c=1}^C L^{\text{vis}}_c = L^{\text{vis}}$, the retention quota is distributed across views by a weighting vector $\alpha \in \mathbb{R}^C$ with $\sum_{c=1}^C \alpha_c = 1$. The remain visual tokens kept for view $c$ is
\begin{equation}
\label{eq:remain visual tokens kept for view}
\mathbf{X}^{\text{vis}}_{(c)} = \operatorname{Top-K}\big(\Phi^{(l)}_{(c)}, k_c\big), \quad
k_c = \left\lfloor \alpha_c \cdot k \right\rfloor, 
\end{equation}
where $\Phi^{(l)}_{(c)}$ denotes the importance scores restricted to view $c$. 
For each view $c$, the remain visual tokens could be represented as:
\begin{equation}
\label{eq:remain visual tokens}
\mathbf{X}^{\text{vis}}_{\text{keep}} 
= \bigcup_{c=1}^C 
   \{\, \mathbf{X}^{\text{vis}}_{(c)}[v] \;\mid\; v \in \mathbf{X}^{\text{vis}}_{(c)},\;
   \mathbf{X}^{\text{vis}}_{(c)}[v] \in \mathbb{R}^D \,\}.
\end{equation}
The reduced visual sequence 
% $\mathbf{H}^{(l)}_{\text{vis,keep}}$
$\mathbf{X}^{\text{vis}}_{\text{keep}}$
is then concatenated back with other modalities to form the pruned multimodal sequence:
\begin{equation}
\tilde{\mathbf{X}}^{m} 
= \,\mathbf{X}^{\text{[BOS]}} \oplus \mathbf{X}^{\text{vis}}_{\text{keep}} \oplus \mathbf{X}^{\text{prop}} \oplus \mathbf{X}^{\text{txt}} \oplus \mathbf{X}^{\text{act}} \oplus \mathbf{X}^{\text{[EOS]}},
\end{equation}
where $\tilde{\mathbf{X}}^{m}$ denotes the dynamically reduced input propagated to the LLM module.

\subsection{Action-Aware Dynamic Strategy}
\label{secsec:Dynamic}
While static pruning
can effectively identify task-relevant visual tokens, not every stage of a manipulation task is suitable for relying solely on the pruned set. In particular, missing fine-grained visual details may cause failures in operations such as moving, swiping, or aligning objects. The accumulation of such local errors can easily propagate and ultimately cause the entire task to fail. To address this limitation, we introduce a dynamic visual strategy that adapts the pruning decision to the robot’s motion state across different task phases, guided by the end-effector (EEF) trajectory within each action chunk, which focus on capturing both translational displacement and rotational motion.

\paragraph{Windowed trajectory and actions.}
Assume $i$ indexes windows and $u$ indexes steps within a window; $b_i$ and $e_i$ denote the start and end timesteps, and $\omega$ matches the OFT action placeholder length.
We treat each decoded action chunk as a temporal window $[b_i, e_i]$ of length $\omega=e_i-b_i+1$ with actions
$\mathbf{A}_i^{\,c}=[\,\mathbf{a}^{\,c}_{i,1};\ldots;\mathbf{a}^{\,c}_{i,\omega}\,]\in\mathbb{R}^{\omega\times 7}$,
where $\mathbf{a}^{\,c}_{i,u}=[\,\Delta x_{i,u},\Delta y_{i,u},\Delta z_{i,u},\Delta\phi_{i,u},\Delta\theta_{i,u},\Delta\psi_{i,u},g_{i,u}\,]^\top$ collects per-step translational and rotational increments and the gripper command.

\begin{definition}[Windowed FK for EEF Position]
\label{def:fk_window}
Let $T_t\in SE(3)$ be the EEF pose at time $t$, and $\pi:SE(3)\!\to\!\mathbb{R}^3$ extract its translation.
With body-frame (right-multiply) composition,
\begin{equation}
\label{eq:fk_window_compact}
\mathbf{p}_{\,b_i+u}\ \triangleq\
\pi\!\left(
T_{b_i}\,\prod_{k=1}^{u}
\begin{bmatrix}
R_{i,k} & \mathbf{v}_{i,k}\\
\mathbf{0}^\top & 1
\end{bmatrix}
\right),\qquad
T_{b_i+u}\ \triangleq\
T_{b_i+u-1}
\begin{bmatrix}
R_{i,u} & \mathbf{v}_{i,u}\\
\mathbf{0}^\top & 1
\end{bmatrix},
\end{equation}
where $\mathbf{v}_{i,u}=[\Delta x_{i,u},\Delta y_{i,u},\Delta z_{i,u}]^\top$ and
$R_{i,u}=R_x(\Delta\phi_{i,u})\,R_y(\Delta\theta_{i,u})\,R_z(\Delta\psi_{i,u})$.
\end{definition}
 The rotation order $R_xR_yR_z$ follows our implementation; if a world-frame (left-multiply) update or a different Euler order is used, the composition should be adjusted accordingly. Definition~\ref{def:fk_window} therefore makes $\mathbf{p}_t=\pi(T_t)$ an explicit function of the full 7-DoF action sequence within each window.

\paragraph{Windowed trajectory distance.}
Given $\mathbf{p}_t$ from Def.~\ref{def:fk_window}, we quantify motion per window either by the Euclidean displacement:
\begin{equation}
\label{eq:cumulative_arc_distance}
\delta_i=\sum_{t=b_i}^{e_i-1}\lVert \mathbf{p}_{t+1}-\mathbf{p}_t\rVert_2.
\end{equation}
The windowed trajectory distance aims to yield a scalar $\delta_i$ that captures overall motion magnitude and subsequently drives our pruning decision rule.

\textbf{Dynamic decision function.}
Given a windowed trajectory distance, we define a binary state variable $s_i \in \{0,1\}$, where $s_i = 0$ corresponds to the \emph{full-vision} state (w/o pruning) and $s_i = 1$ to the \emph{pruned} state (cross-attention pruning). 
therefore next state could be determined by
\begin{equation}
\label{eq:next state}
s_{i+1} = f(\delta_i) = 
\begin{cases}
1, & \delta_i \ge \bar{\delta}_i, \\
0, & \delta_i < \bar{\delta}_i;
\end{cases}
\qquad
\bar{\delta}_i = \frac{1}{i} \sum_{j=1}^i \delta_j. 
\end{equation}
The dynamic switch to $s_{i+1}$ adapts to the global motion scale of the task, enabling pruning during periods of high activity while retaining full vision during fine-grained phases. An alternative is the \emph{adjacent-extrema function}, which sets thresholds based on the extrema of the most recent $\tau$ windows:
\begin{equation}
\begin{aligned}
    U^{(i)} &= \max\{\delta_{i-\tau+1}, \dots, \delta_i\}, \\
    V^{(i)} &= \min\{\delta_{i-\tau+1}, \dots, \delta_i\}
\end{aligned}
\end{equation}
with the update rule
\begin{equation}
\label{eq:next state2}
s_{i+1} =
\begin{cases}
1, & \delta_i \ge U^{(i)}, \\
0, & \delta_i \le V^{(i)}, \\
s_i, & V^{(i)} < \delta_i < U^{(i)}.
\end{cases}
\end{equation}
This design aims to respond quickly to local motion variations, capturing abrupt efficient shifts between coarse and delicate manipulations.
When the robot exhibits large-amplitude motions, pruning is activated to suppress redundant visual input and reduce computation. When motion amplitude decreases, signaling the onset of fine manipulation, pruning is disabled to preserve the complete visual context necessary for accurate control.

\subsection{Theoretical Analysis of Computational Complexity}
We consider the computational cost of the Transformer stack in $f_{\text{LLM}}$ with respect to the sequence length before and after pruning.
Let $S$ denote the token length of the full multi-modal sequence in Eq.~\ref{eq:pre:emb}, $D$ the hidden dimension, and $M$ the intermediate dimension of the feed-forward network (SwiGLU).
The total number of vision tokens before pruning is $L_{\text{vis}}$, while $k=\lfloor \rho \cdot L_{\text{vis}} \rfloor$ tokens are kept after the text-driven selection in Eq.~\ref{eq:keep embeddings}.
We denote by $T$ the number of forwards in a complete task execution (Eq.~\ref{eq:oft_parallel}), and by $\gamma\in[0,1]$ the proportion of forwards executed in the pruned state (Sec.~\ref{secsec:Dynamic}).
Let $H$ be the number of Transformer layers in $f_{\text{LLM}}$.

For one Transformer layer, the FLOPs are approximated by
\begin{equation}
F(S;D,M) \approx 2 S^2 D + 4 S D^2 + 6 S D M,
\end{equation}
corresponding to attention, projections, and MLP. A baseline forward of the $H$-layer LLM therefore costs
\begin{equation}
F_{\text{base}} = H \cdot F(S;D,M).
\end{equation}

In ADP, pruning happens \emph{before} the LLM at the embedding stage: visual tokens are ranked and reduced using Eq.~\ref{eq:attention score}–\ref{eq:keep embeddings}, yielding the shorter sequence
\begin{equation}
S' = 1 + k + L_{\text{prop}} + L_{\text{txt}} + L_{\text{act}} + 1.
\end{equation}
The scoring overhead uses lightweight projections and a similarity matrix between text and vision embeddings,
\begin{equation}
F_{\text{score}} = 2 L_{\text{txt}} D^2 + 2 L_{\text{vis}} D^2 + 2 N^h L_{\text{txt}} L_{\text{vis}} d.
\end{equation}
Assume pruning and scoring are performed once per forward \emph{before} any Transformer layer, so all $H$ layers operate on $S'$, the per-layer cost follows $F(S;D,M)\!\approx\!2S^2D+4SD^2+6SDM$, and $D=N^h d$, 
\begin{equation}
\Delta F_{\text{ADP}} = F_{\text{base}} - F_{\text{ADP}}, \quad
% F_{\text{post}} = H \cdot F(S';D,M), \quad 
F_{\text{ADP}} = F_{\text{score}} + H \cdot F(S';D,M), 
\end{equation}
where $k=\lfloor \rho\,L_{\text{vis}}\rfloor$, $S'$ is as above with $L_{\text{prop}},L_{\text{txt}},L_{\text{act}}$ the proprioception/text/action lengths, $N^h$ is the number of attention heads of size $d$, and $F_{\text{base}}=H\,F(S;D,M)$.
Over an episode of $T$ forwards, the expected complexity under the dynamic strategy is
\begin{equation}
\mathbb{E}\!\left[\mathcal{F}_{\mathrm{episode}}\right]
= T \big(\gamma\,F_{\text{ADP}} + (1-\gamma)\,F_{\text{base}}\big),
\end{equation}
with expected savings
\begin{equation}
\mathbb{E}\!\left[\Delta \mathcal{F}_{\mathrm{episode}}\right]
= T\,\gamma\,\Delta F_{\text{ADP}}.
\end{equation}
The pruning is applied at the embedding stage prior to $f_{\text{LLM}}$, so the reduced length $S'$ benefits all $H$ layers uniformly, and the dynamic rule (Eq.~\ref{eq:fk_window_compact}–\ref{eq:next state}) controls how often the pruned path is used across action windows.

\definecolor{customblue}{rgb}{0,0,1}
\colorlet{custombluealpha}{customblue!60}

%% file: sec/4_experiment.tex
\section{EXPERIMENTS}
We conduct experiments across LIBERO simulation and real-robot tasks under standardized settings, comparing against strong baselines and conducting targeted ablations to assess success rates, compute/latency, and the contributions of dynamic scheduling and layer-wise pruning.

\begin{table*}[t]
\centering
\caption{\textbf{Results on the LIBERO Benchmark.} \textbf{TR}: Training-Free; \textbf{AR}: Auto-Regressive; \textbf{PD}: Parallel Decoding. \textbf{Ratio}: The retain tokens / Full tokens.
}
\label{fig:SIMULATION EXPERIMENTS}
\resizebox{\textwidth}{!}{
\begin{tabular}{l|ccc|ccccc|cc}
\toprule
\textbf{Method} & \textbf{TR} & \textbf{Decoding} & \textbf{CKPT} & \textbf{Spatial} & \textbf{Object} & \textbf{Goal} & \textbf{Long} & \textbf{Average} & \textbf{FLOPs}$\downarrow$ & \textbf{Speedup}$\uparrow$\\
\midrule
% --------- Literature block first (no citations) ----------
OpenVLA~\cite{kim2024openvla} & - & AR & OpenVLA (7B) & 84.7\% & 88.4\% & 79.2\% & 53.7\% & 76.5\% & - & - \\
SparseVLM~\cite{zhang2024sparsevlm} & \textcolor{green}{\cmark} & AR & OpenVLA (7B) & 79.8\% & 67.0\% & 72.6\% & 39.4\% & 64.7\% & - & - \\
FastV~\cite{chen2024image} & \textcolor{green}{\cmark} & AR & OpenVLA (7B) & 83.4\% & 84.0\% & 74.2\% & 51.6\% & 73.3\% & - & - \\
VLA-Cache~\cite{xu2025vla} & \textcolor{green}{\cmark} & AR & OpenVLA (7B) & 83.8\% & 85.8\% & 76.4\% & 52.8\% & 74.7\% & - & - \\
FlashVLA~\cite{tan2025think} & \textcolor{green}{\cmark} & AR & OpenVLA (7B) & 84.2\% & 86.4\% & 75.4\% & 51.4\% & 74.4\% & - & - \\
SP-VLA~\cite{li2025sp} & \textcolor{green}{\cmark} & AR & OpenVLA (7B) & 75.4\% & 85.6\% & 84.4\% & 54.2\% & 74.9\% & - & - \\
WorldVLA~\cite{cen2025worldvla} & - & AR & Chameleon (7B) & 85.6\% & 89.0\% & 82.6\% & 59.0\% & 79.1\% & - & - \\
WorldVLA*~\cite{cen2025worldvla} & - & AR & Chameleon (7B) & 87.6\% & 96.2\% & 83.4\% & 60.0\% & 81.8\% & - & - \\
NORA~\cite{hung2025nora} & - & AR & Qwen-VL (3B) & 85.6\% & 87.8\% & 77.0\% & 45.0\% & 73.9\% & - & - \\
SmolVLA~\cite{shukor2025smolvla} & - & AR & SmolVLM (2.25B) & 93.0\% & 94.0\% & 91.0\% & 77.0\% & 88.8\% & - & - \\
CogACT~\cite{li2024cogact} & - & FM & CogVLM (7B) & 97.2\% & 98.0\% & 90.2\% & 88.8\% & 93.6\% & - & - \\
CSP~\cite{pei2024cross} & \textcolor{green}{\cmark} & PD & OFT (7B) & 84.7\% & 82.2\% & 77.1\% & 74.3\% & 79.6\% & - & - \\
NORA-Long~\cite{hung2025nora} & - & PD & Qwen-VL (3B) & 92.2\% & 95.4\% & 89.4\% & 74.6\% & 87.9\% & - & - \\
% --------- Your methods block ----------
OpenVLA-OFT~\cite{kim2025fine} & - & PD & OFT (7B) & 98.6\% & 98.2\% & 96.6\% & 94.8\% & 97.1\% & 7.91 & 1.00× \\
FastV(+OFT)~\cite{chen2024image} & \textcolor{green}{\cmark} & PD & OFT (7B) & 96.8\% & 81.0\% & 96.4\% & 73.0\% & 86.8\% & 6.37 & 1.24× \\
\midrule
\rowcolor{lavender}\textbf{VLA-ADP (Ratio=30\%)} & 
\textcolor{green}{\cmark} & PD & OFT (7B) & 97.6\% & \textbf{98.4\%} & \textbf{97.4\%} & 84.2\% & 94.4\% & 5.85 & 1.35× \\
\rowcolor{lavender}\textbf{VLA-ADP (Ratio=40\%)} & \textcolor{green}{\cmark} & PD & OFT (7B) & 98.2\% & 97.2\% & 96.6\% & 87.2\% & 94.8\% & 6.14 & 1.29× \\
\rowcolor{lavender}\textbf{VLA-ADP (Ratio=50\%)} & \textcolor{green}{\cmark} & PD & OFT (7B) & \textbf{99.4}\% & 98.0\% & 96.4\% & 91.2\% & \textbf{96.3\%} & 6.43 & 1.23× \\
\rowcolor{lavender}\textbf{VLA-ADP (Ratio=60\%)} & \textcolor{green}{\cmark} & PD & OFT (7B) & 98.8\% & 98.0\% & 95.8\% & \textbf{92.0\%} & 96.2\% & 6.74 & 1.17× \\
\rowcolor{lavender}\textbf{VLA-ADP (Ratio=70\%)} & \textcolor{green}{\cmark} & PD & OFT (7B) & 99.0\% & 98.2\% & 96.8\% & 91.2\% & \textbf{96.3\%} & 7.03 & 1.13× \\
\bottomrule
\end{tabular}
}
% \vspace{-0.1in}
\end{table*}

\begin{figure*}[t]
    \centering
    \includegraphics[width=0.99\linewidth]{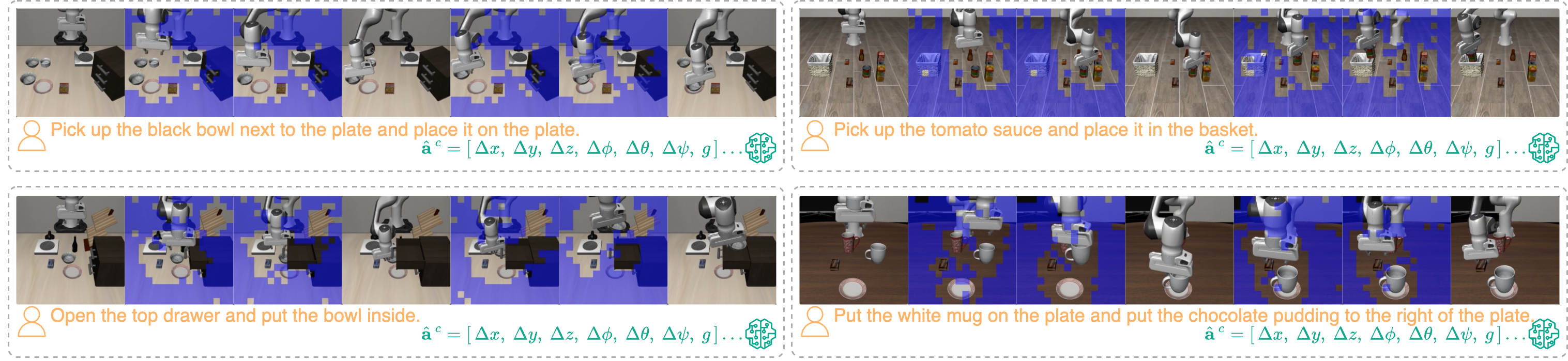}
    \caption{Visualisation of our method on representative examples of the four LIBERO task types (Spatial, Object, Goal, Long). Blue masks \textcolor{custombluealpha}{\rule{7pt}{7pt}} indicate pruned vision tokens. The retained tokens consistently highlight task-relevant objects, validating the Text-driven Anticipatory Pruning. Moreover, full vision tokens are restored at critical phases (\textit{e.g.}, initialisation, grasping, placement), demonstrating the effectiveness of the Task-driven Pruning and Action-Aware Dynamic Strategy.}
    \label{fig:liberoplot}
    \vspace{-0.1in}
\end{figure*}

\subsection{SIMULATION EXPERIMENTS}
\textbf{Experiments setup.} 
For simulation, we use LIBERO \cite{liu2023libero} with four suites (\textit{i.e.}, Spatial, Object, Goal, Long) evaluating spatial understanding, object recognition, goal-directed behaviour, and long-horizon planning. All Libero settings follow OpenVLA-OFT, using its public run scripts. For comparison, we reproduce FastV \cite{chen2024image} on OpenVLA-OFT and run under the same environment. Experiments run on Linux with an NVIDIA RTX 4090. We set the window size to the OpenVLA-OFT chunk size (8) and apply a cold start: the first two windows use full vision. To limit error accumulation, if pruning occurs in three consecutive windows, the next window is forced to full vision. In multi-view pruning, the main\:wrist retention is 4:6. Within the Action-Aware Dynamic Strategy, we use Euclidean displacement and the adjacent-extrema rule; in its third case we deterministically set the state to 1 (instead of inheriting $s_i$) to further reduce FLOPs.

\textbf{Main result.} 
Table \ref{fig:SIMULATION EXPERIMENTS} shows that VLA-ADP achieves a stable accuracy–compute trade-off across the keep ratio on LIBERO. Compared to OpenVLA-OFT, when the keep ratio is 50–70\%, the average success rate slightly decreases ($\leq$ 0.9\%), while the LLM-side inference speed improves by up to 1.23×, and the FLOPs are markedly below the baseline. Further compressing the retention rate to 30–40\% maintains an average SR of 94.4–94.8\%, yet attains 1.29–1.35× speedup. Notably, VLA-ADP achieves a 99.4\% success rate on the Spatial, indicating that VLA-ADP successfully prunes redundant vision tokens while selectively preserving key information in relatively simple spatial manipulation scenarios.
By contrast, Random Dropping (50\%) yields a 1.29× speedup on the LLM side and performs reasonably on the Spatial and Goal, but its success rates on Object and Long are only 73.0\% and 76.2\%, respectively.

\subsection{REAL-WORLD EXPERIMENTS}
\label{sec:REAL-WORLD EXPERIMENTS}

\begin{table*}[t]
\centering
\caption{\textbf{Real-World Experiments (4 tasks).} Task 1 to Task 4 cover the picking, placing and wiping motion, which covering a variety of objects.
}
\resizebox{\textwidth}{!}{
\begin{tabular}{lccccccccc}
\toprule
\textbf{Method}  & \textbf{Decoding} &  \textbf{Task1} & \textbf{Task2} & \textbf{Task3} & \textbf{Task4} & \textbf{Average} &  \textbf{Latency}$\downarrow$ & \textbf{Speedup}$\uparrow$ \\
\midrule
OpenVLA-OFT (base) & PD &  83.3\% & \textbf{93.3\%} & 86.7\% & 80.0\% & 85.8\% &  76.9 & 1.00 \\
\textbf{VLA-ADP} (ours) & PD &  \textbf{90.0\%} & 90.0\% & \textbf{90.0\%} & \textbf{83.3\%} & \textbf{88.3\%}  & \textbf{51.8} & \textbf{1.49}$\times$ \\
\bottomrule
\end{tabular}
}
\label{tab:real}
% \vspace{-0.1in}
\end{table*}
\begin{figure}[!t]
    \centering
    \includegraphics[width=1\linewidth]{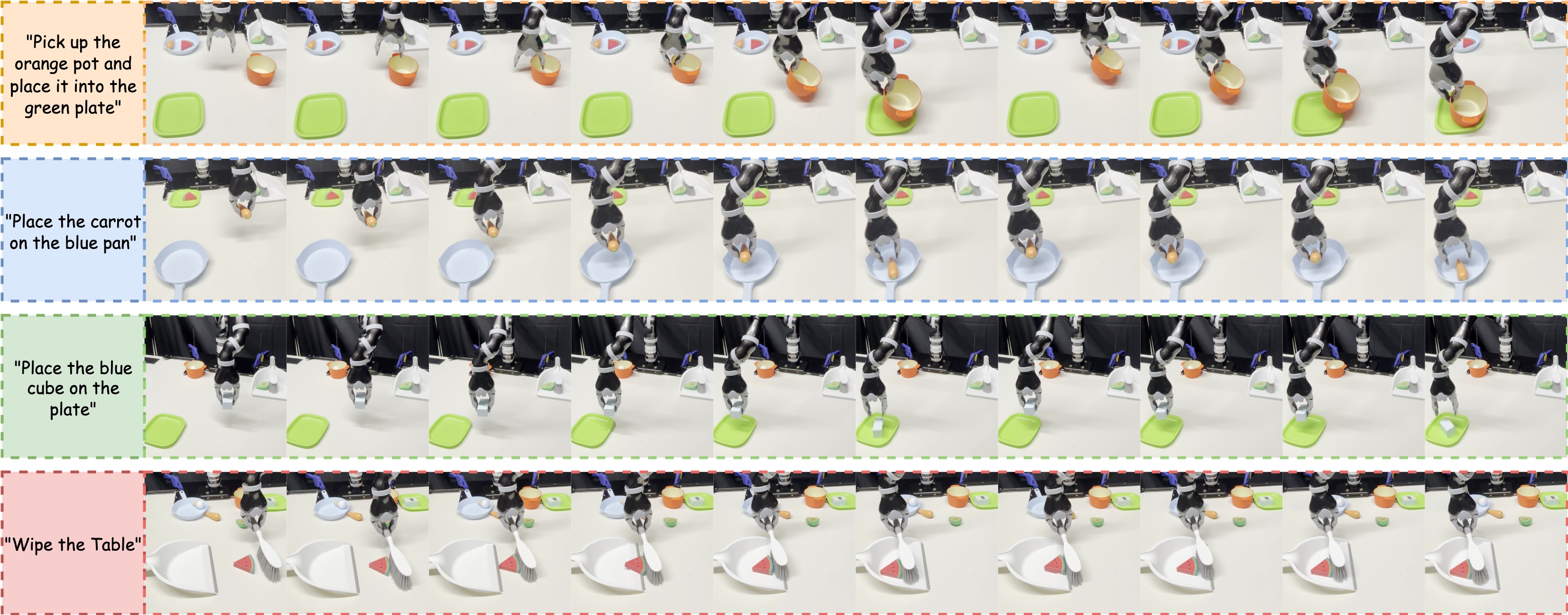}
    \caption{\textbf{Real world experiments}. We conduct the experiments on Jaco2 Real-world Platform.}
    \label{fig:real}
    \vspace{-0.1in}
\end{figure}

\textbf{Experiment setup.}
We evaluate on four real tasks executed on a physical robot:
% \begin{enumerate}
%     \item pick up the orange pot and place it in the green plate.
%     \item place the blue cube on the plate.
%     \item place the carrot on the blue pan.
%     \item wipe the table.
% \end{enumerate}
\emph{Task1}: pick up the orange pot and place it in the green plate;
\emph{Task2}: place the blue cube on the plate;
\emph{Task3}: place the carrot on the blue pan;
\emph{Task4}: wipe the table.
All runtime settings mirror the simulation: Linux workstation with an NVIDIA RTX~4090, parallel decoding (PD) and window size equal to OpenVLA-OFT chunk size (8) with a two-window cold start (full vision). To curb error accumulation, if pruning is enabled for three consecutive windows, the next window uses full vision. Within the Action-Aware Dynamic Strategy we use Euclidean displacement and the adjacent-extrema rule, and in the third case deterministically set the state to 1 to further reduce FLOPs.

\textbf{Main result.}
Table~\ref{tab:real} summarizes the outcomes. VLA-ADP improves the average success rate from \mbox{85.8\%} to \mbox{88.3\%} while reducing latency from \mbox{76.9} to \mbox{51.8} (baseline/ours $= \mathbf{1.49}\times$). Per task, VLA-ADP outperforms the baseline on Task1/3/4 (\mbox{90.0\%}, \mbox{90.0\%}, \mbox{83.3\%}) and is still competitive on Task2 (90.0\% vs.\ 93.3\%). These results indicate that our dynamic pruning maintains or improves real-world success while yielding substantial speedup.

\section{Ablation Study}
\label{sec:Ablation Study}

We conduct ablations on each component of our method: first removing the dynamic strategy to test ADP (\emph{w/o Action-aware Dynamic}); then ablating the pruning module (\emph{w/o Text-driven Pruning}) with different weight selections for relevance scoring.

\begin{table*}[th]
    \centering
    \scriptsize
    \caption{\textbf{Ablation Study on each component of our design.} \textbf{(a)} Ablation on dynamic strategies (\(f\)). \textbf{(b)} Ablation on pruning based on weights retrievaled from layer $l$. Metrics are SR (\%) and FLOPs. 
    % (GFLOPs/forward).
    }
    \renewcommand{\arraystretch}{1.05}
    \begin{subtable}[t]{0.55\textwidth}
        \centering
        \setlength{\tabcolsep}{3.6pt}
        \begin{tabular}{@{}l c c c c ccc@{}}
        \toprule
        \multicolumn{1}{c}{\multirow{2}{*}[-0.2ex]{\(f\)}}
          & \multicolumn{1}{c}{Spatial}
          & \multicolumn{1}{c}{Object}
          & \multicolumn{1}{c}{Goal}
          & \multicolumn{1}{c}{Long}
          & \multicolumn{3}{c}{Avg} \\
        \cmidrule(lr){2-2}\cmidrule(lr){3-3}\cmidrule(lr){4-4}\cmidrule(lr){5-5}\cmidrule(lr){6-8}
          & SR$\uparrow$ & SR$\uparrow$ & SR$\uparrow$ & SR$\uparrow$ & SR$\uparrow$ & $\rho^{avg}$ & FLOPs$\downarrow$ \\
        \midrule
        % Base & 98.2\% & 88.0\% & 96.4\% & 91.2\% & 93.45\% & 0.25 & 6.23 \\
        ADP   & \textbf{99.4\%} & \textbf{98.0\%} & \textbf{96.4\%} & \textbf{91.2\%} & \textbf{96.3\%} & \textbf{0.22} & \textbf{6.43} \\
        - w/o D & 98.2\% & 88.0\% & 96.4\% & 91.2\% & 93.45\% & 0.25 & 6.23 \\
        - w/o D + PS  & 95.0\% & 81.4\% & 96.2\% & 87.0\% & 89.9\%  & 0.50 & 4.55 \\
        \bottomrule
        \end{tabular}
        \caption{}
        \label{tab:ablation_dynamic}
        \vspace{-3mm}
    \end{subtable}%  <-- keep this percent to suppress inter-column space
    \hfill
    \begin{subtable}[t]{0.42\textwidth}
        \centering
        \setlength{\tabcolsep}{3.6pt}
        \begin{tabular}{@{}c c c c c cc@{}}
        \toprule
        \multicolumn{1}{c}{\multirow{2}{*}[-0.2ex]{\(l\)}}
          & \multicolumn{1}{c}{Spatial}
          & \multicolumn{1}{c}{Object}
          & \multicolumn{1}{c}{Goal}
          & \multicolumn{1}{c}{Long}
          & \multicolumn{2}{c}{Avg} \\
        \cmidrule(lr){2-2}\cmidrule(lr){3-3}\cmidrule(lr){4-4}\cmidrule(lr){5-5}\cmidrule(lr){6-7}
          & SR$\uparrow$ & SR$\uparrow$ & SR$\uparrow$ & SR$\uparrow$ & SR$\uparrow$ & FLOPs$\downarrow$ \\
        \midrule
        0 & \textbf{99.4\%} & \textbf{98.0\%} & \textbf{96.4\%} & \textbf{91.2\%} & \textbf{96.3\%} & \textbf{6.43} \\
        1 & 98.2\% & 97.6\% & 96.2\% & 89.8\% & 95.5\% & 6.57 \\
        4 & 98.0\% & 97.8\% & 96.4\% & 91.2\% & 95.8\% & 6.89 \\
        % 8 & 98.6\% & \textbf{98.0\%} & \textbf{96.6\%} & 89.6\% & 95.7\% & 7.31 \\
        \bottomrule
        \end{tabular}
        \caption{}
        \label{tab:ablation_layer_rho}
        \vspace{-3mm}
    \end{subtable}
    \vspace{-0.1in}
\end{table*}

\begin{figure*}[ht]
    \centering
    \begin{subfigure}{0.32\textwidth}
        \centering
        \includegraphics[width=0.48\textwidth]{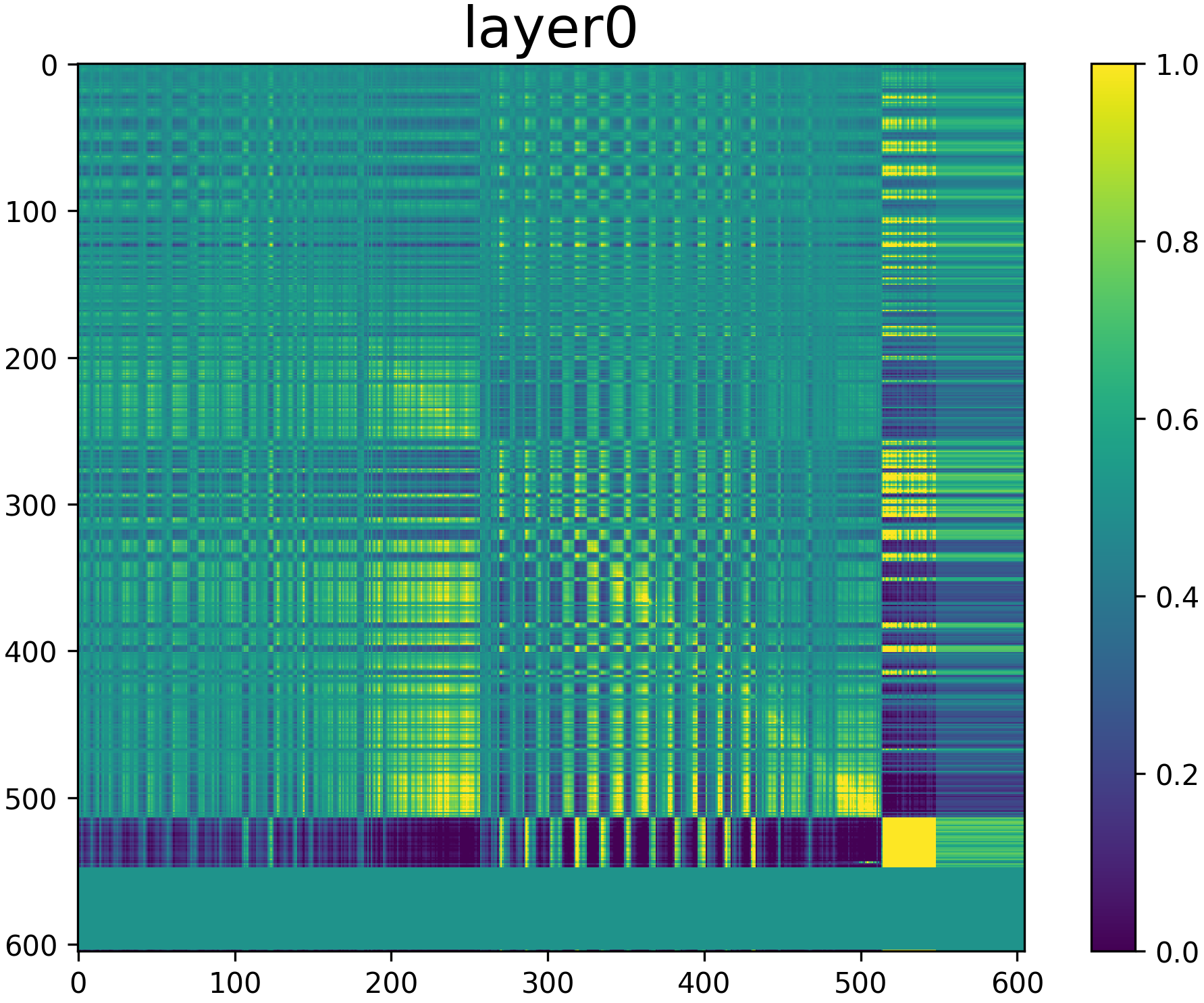}
        \includegraphics[width=0.48\textwidth]{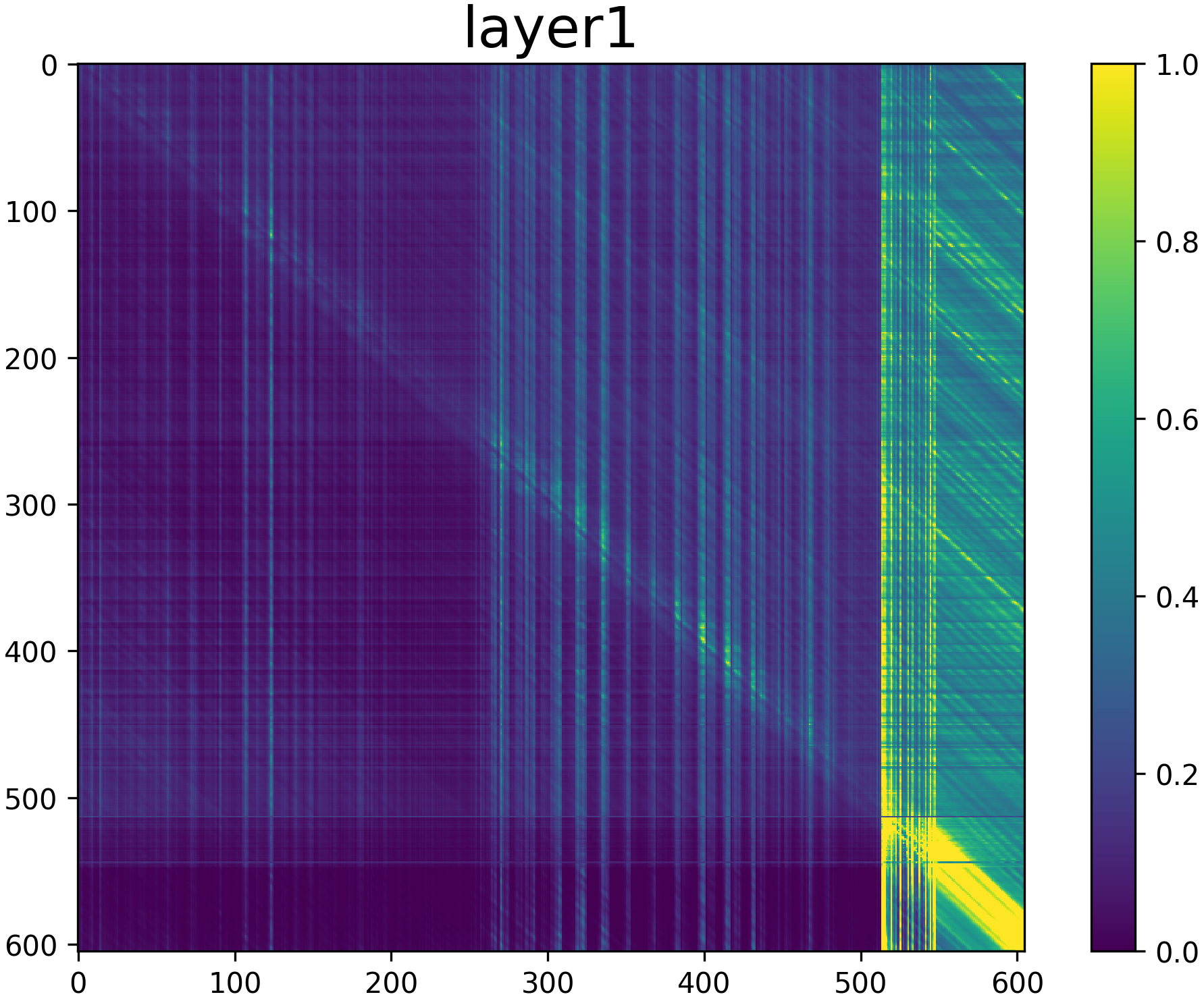}
        \includegraphics[width=0.48\textwidth]{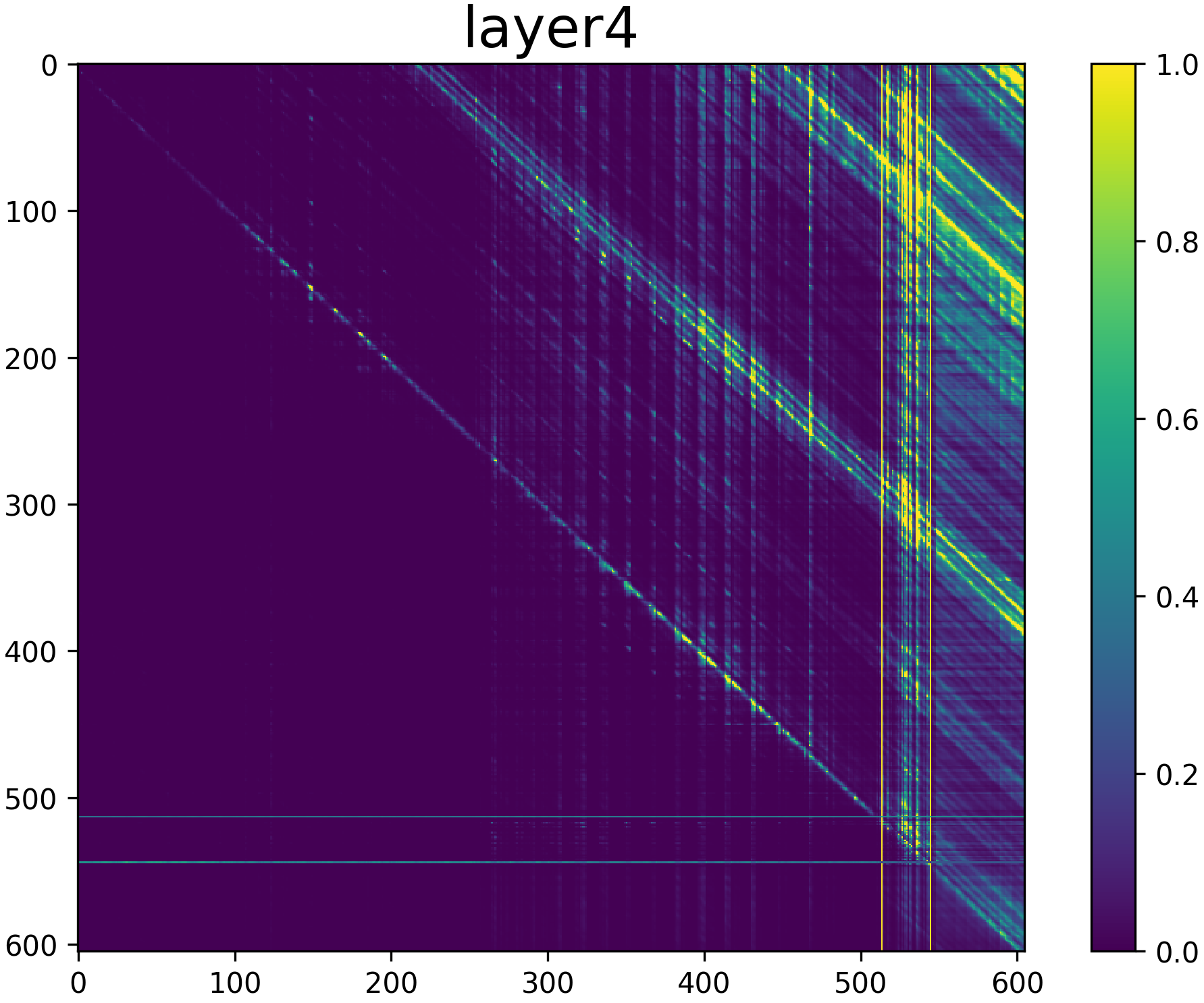}
        \includegraphics[width=0.48\textwidth]{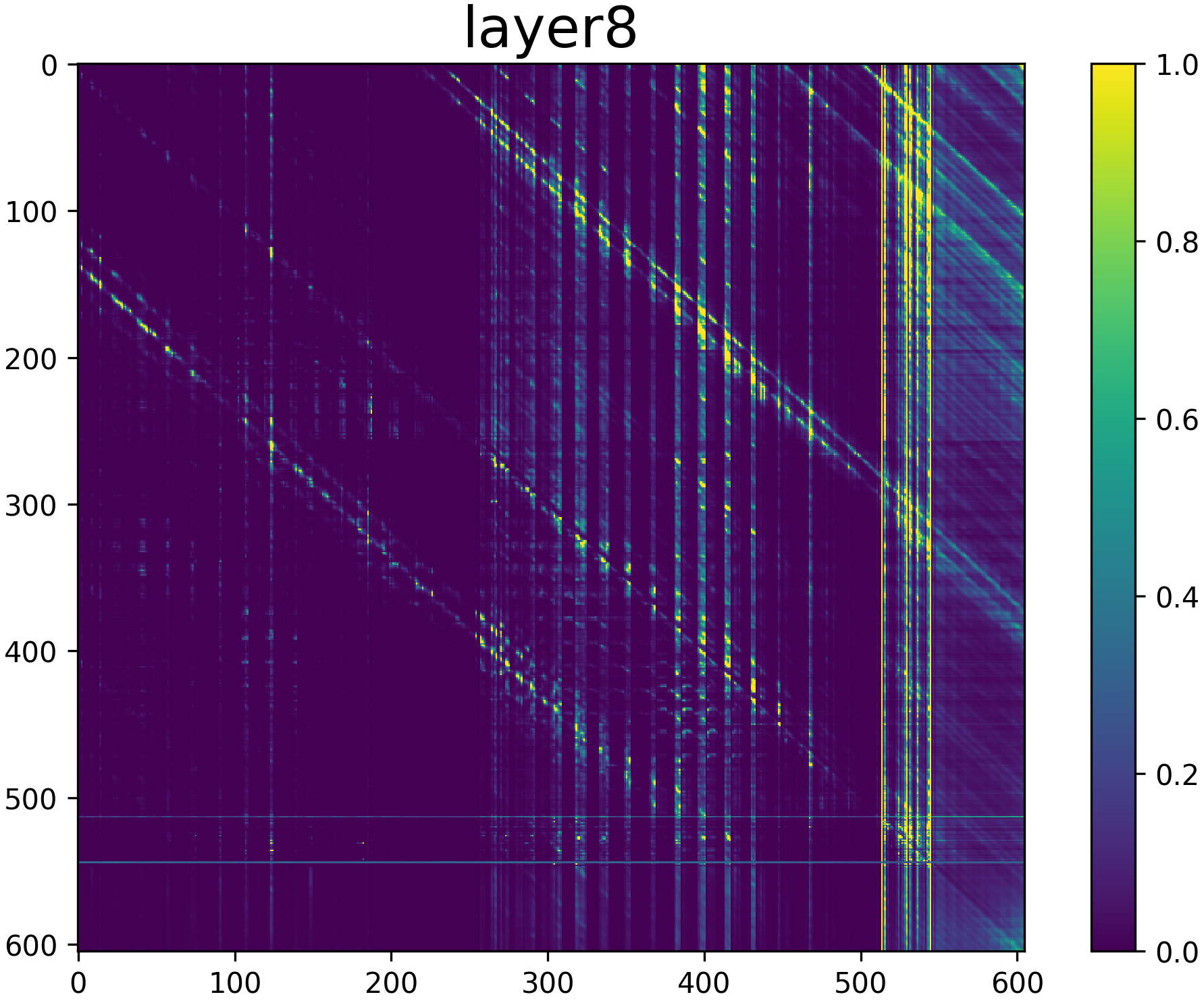}
        \caption{Original Attention Maps}
        \label{fig:smss}
    \end{subfigure}
    \hfill
    \begin{tikzpicture}[remember picture,overlay]
        \draw[dashed, thick] (-0.15, 0.55) -- (-0.15,4.5);
    \end{tikzpicture}
    \begin{subfigure}{0.32\textwidth}
        \centering
        \includegraphics[width=0.48\textwidth]{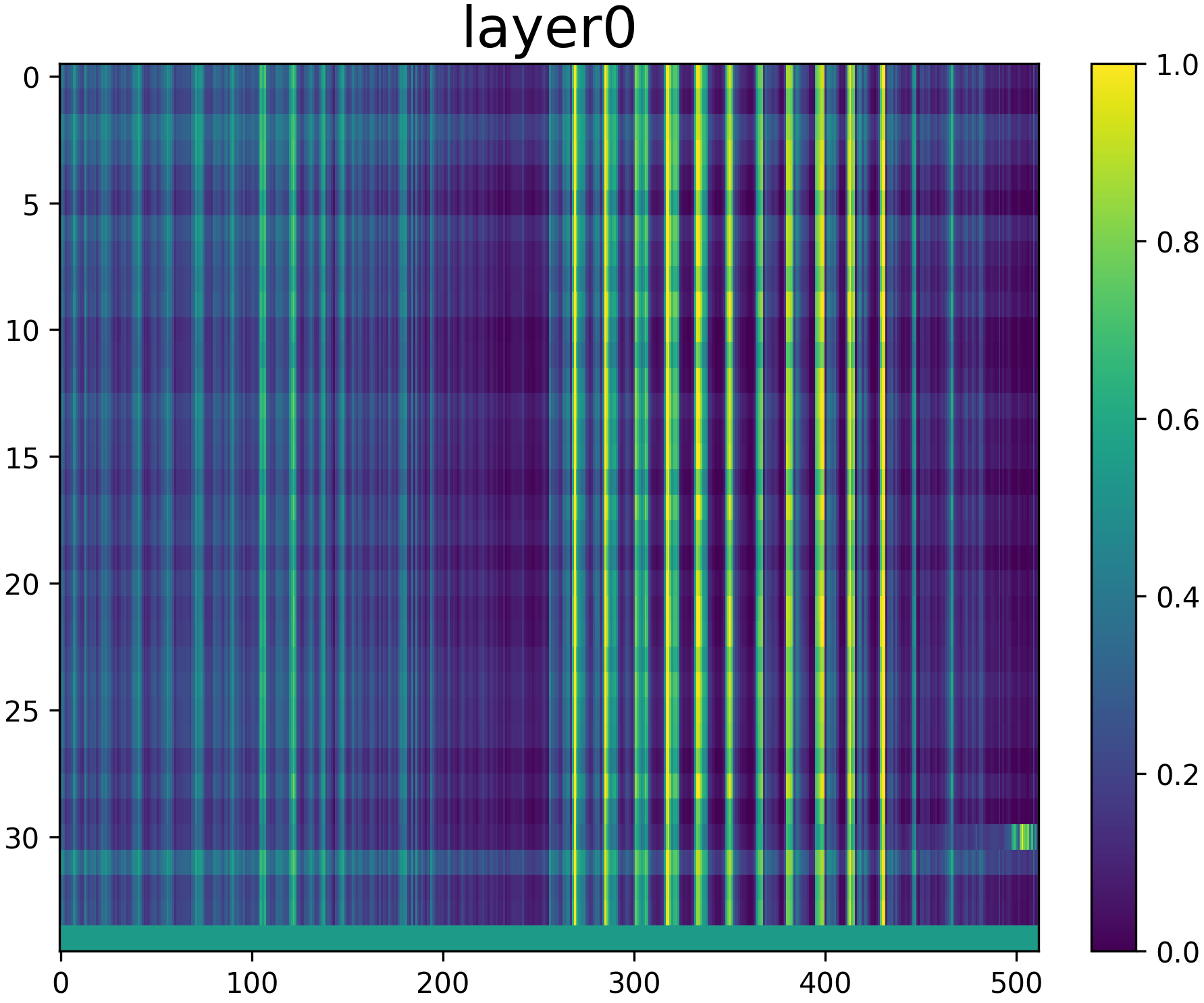}
        \includegraphics[width=0.48\textwidth]{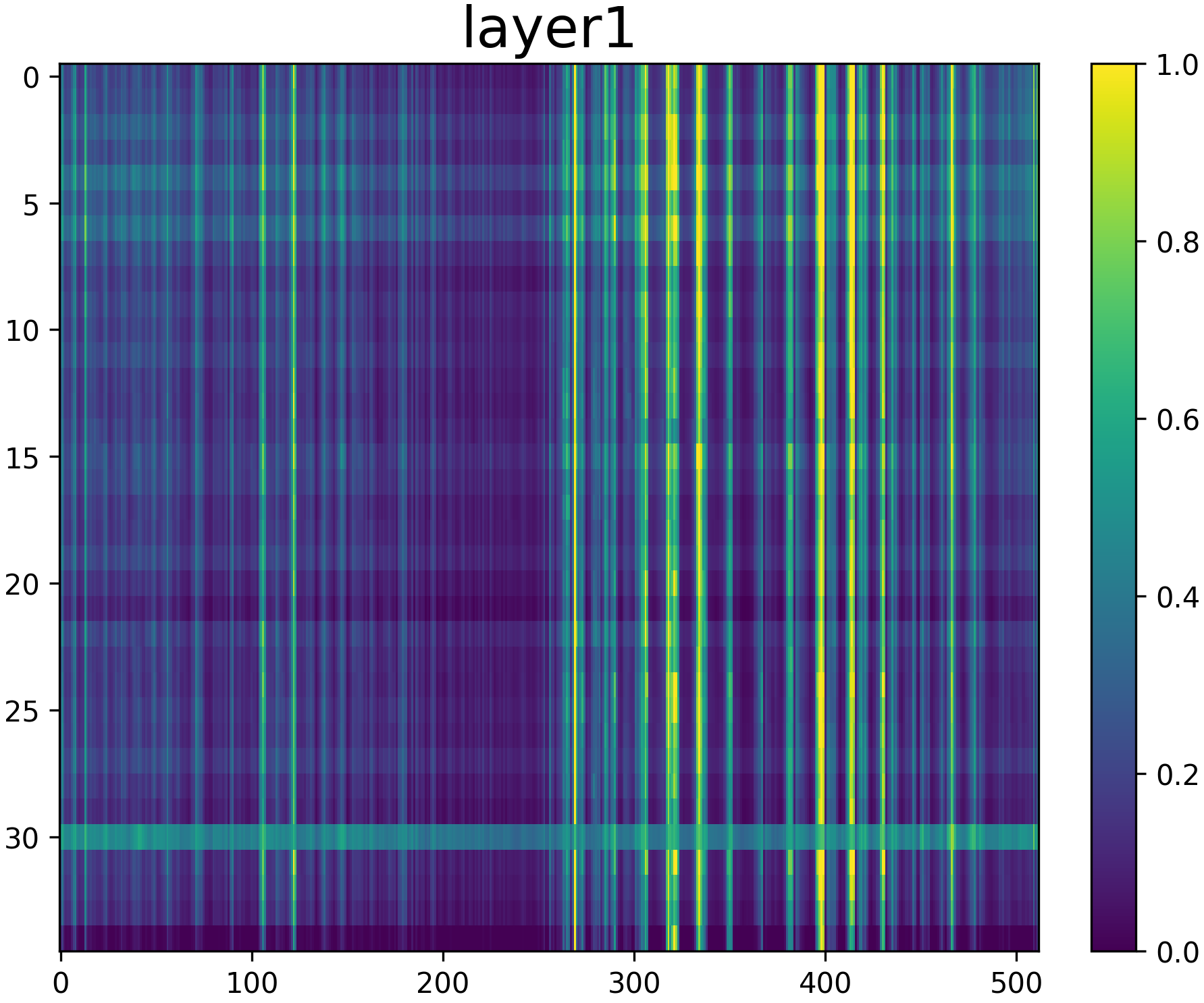}
        \includegraphics[width=0.48\textwidth]{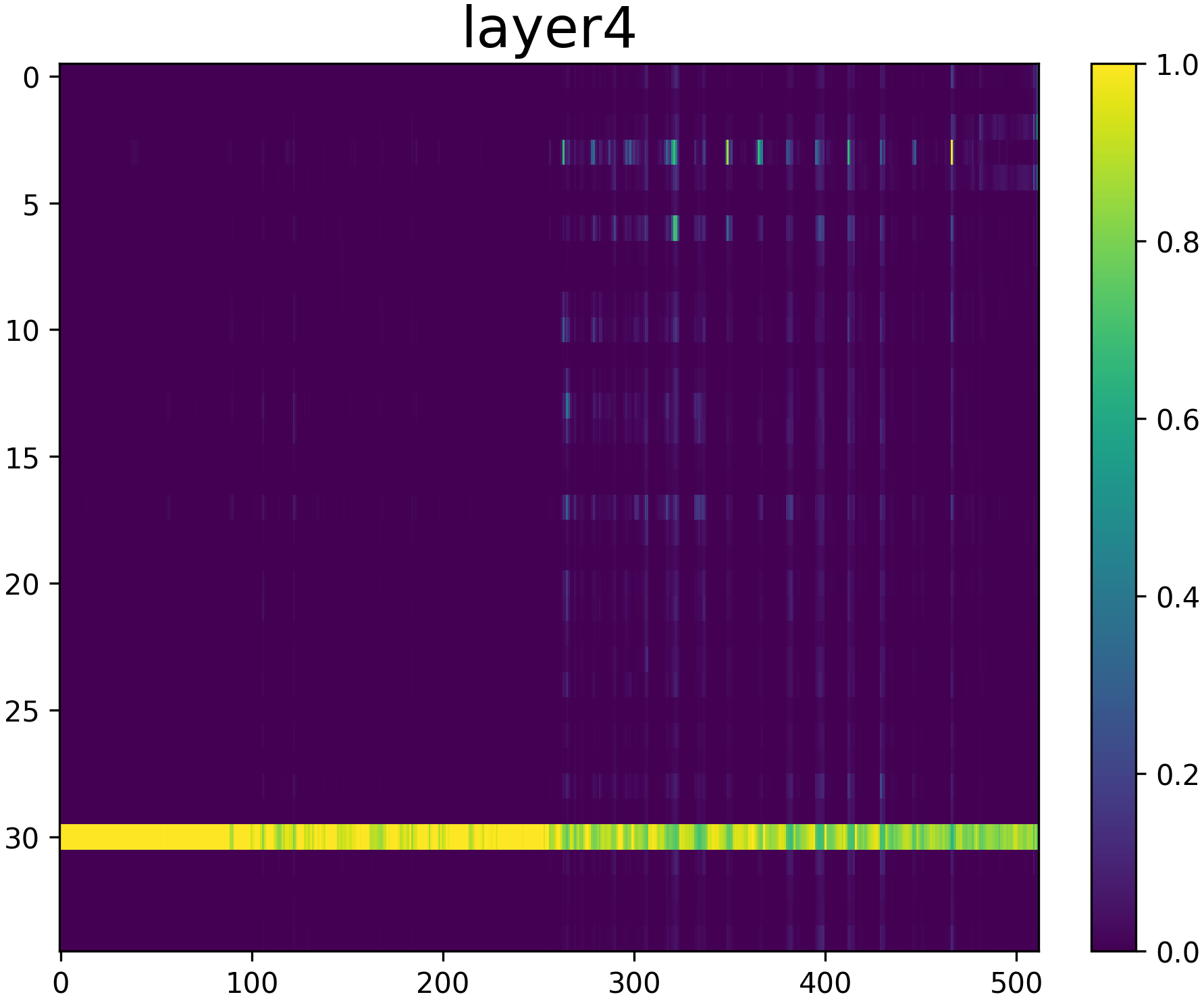}
        \includegraphics[width=0.48\textwidth]{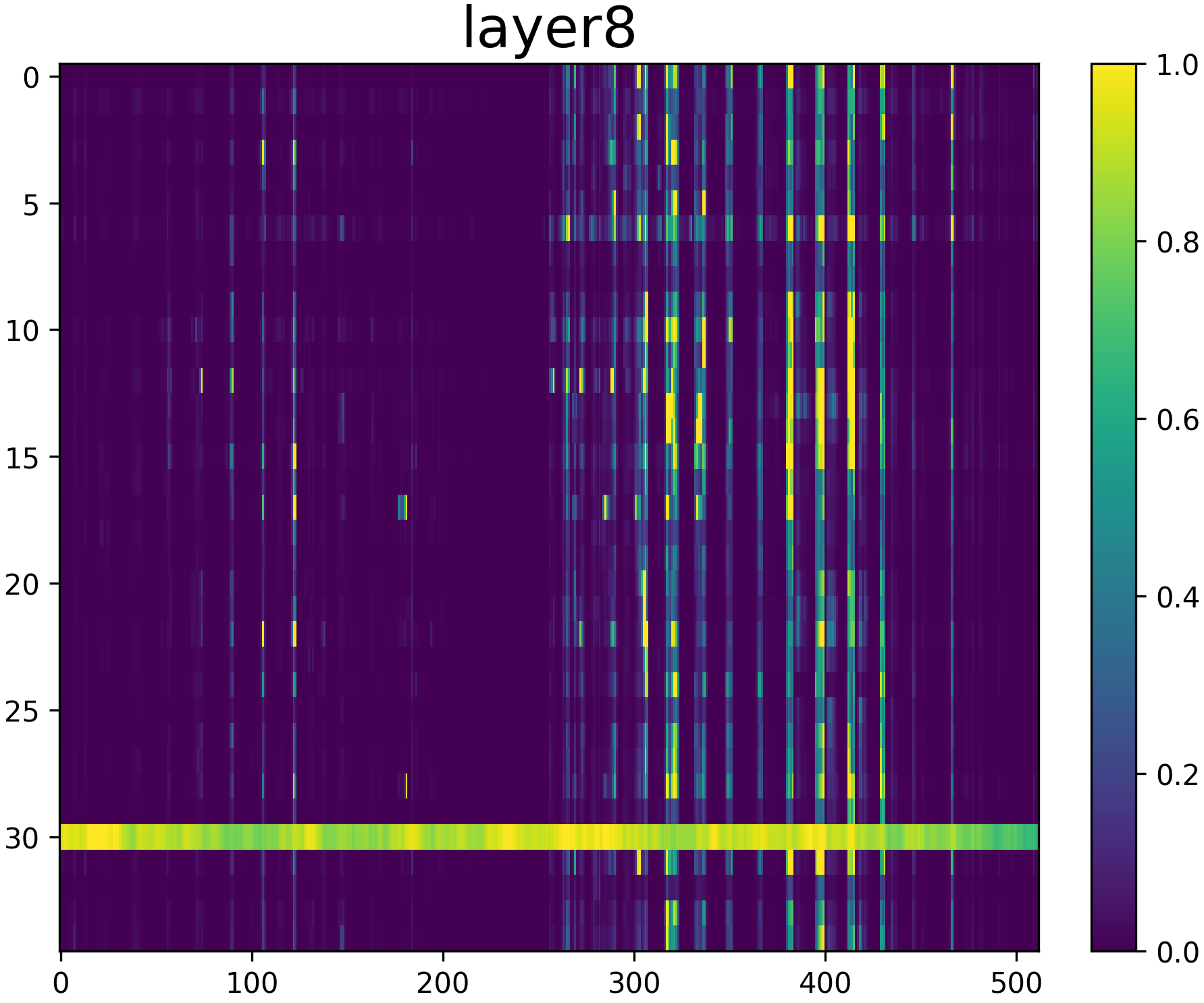}
        \caption{Pruning Attention Maps.}
        \label{fig:smtv}
    \end{subfigure}
    \hspace{-0.015\textwidth}
    \begin{tikzpicture}[remember picture,overlay]
        \draw[dashed, thick] (0.25, 0.55) -- (0.25,4.5);
    \end{tikzpicture}
    \begin{subfigure}{0.32\textwidth}
        \centering
        \includegraphics[width=0.405\textwidth]{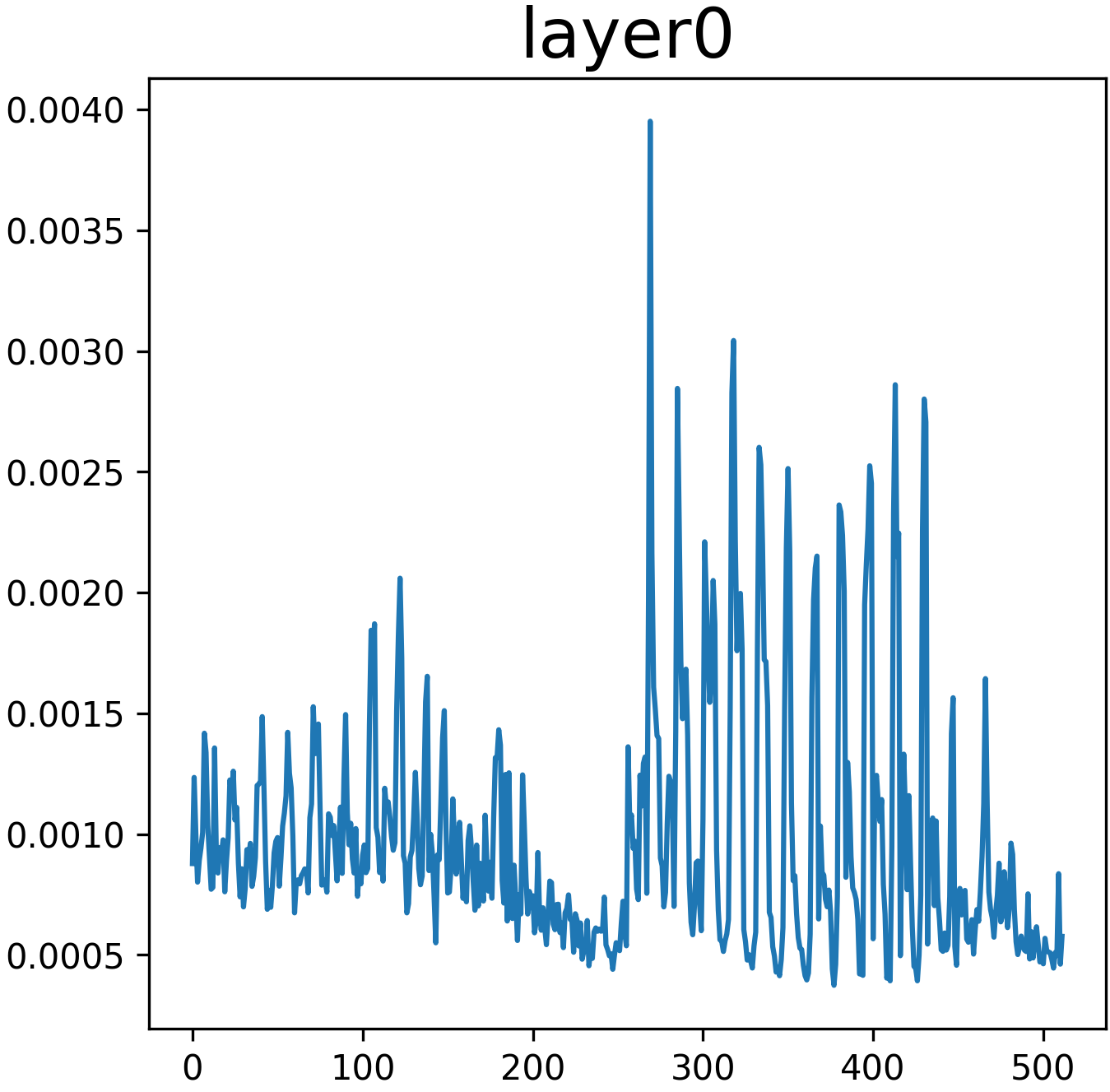}
        \includegraphics[width=0.405\textwidth]{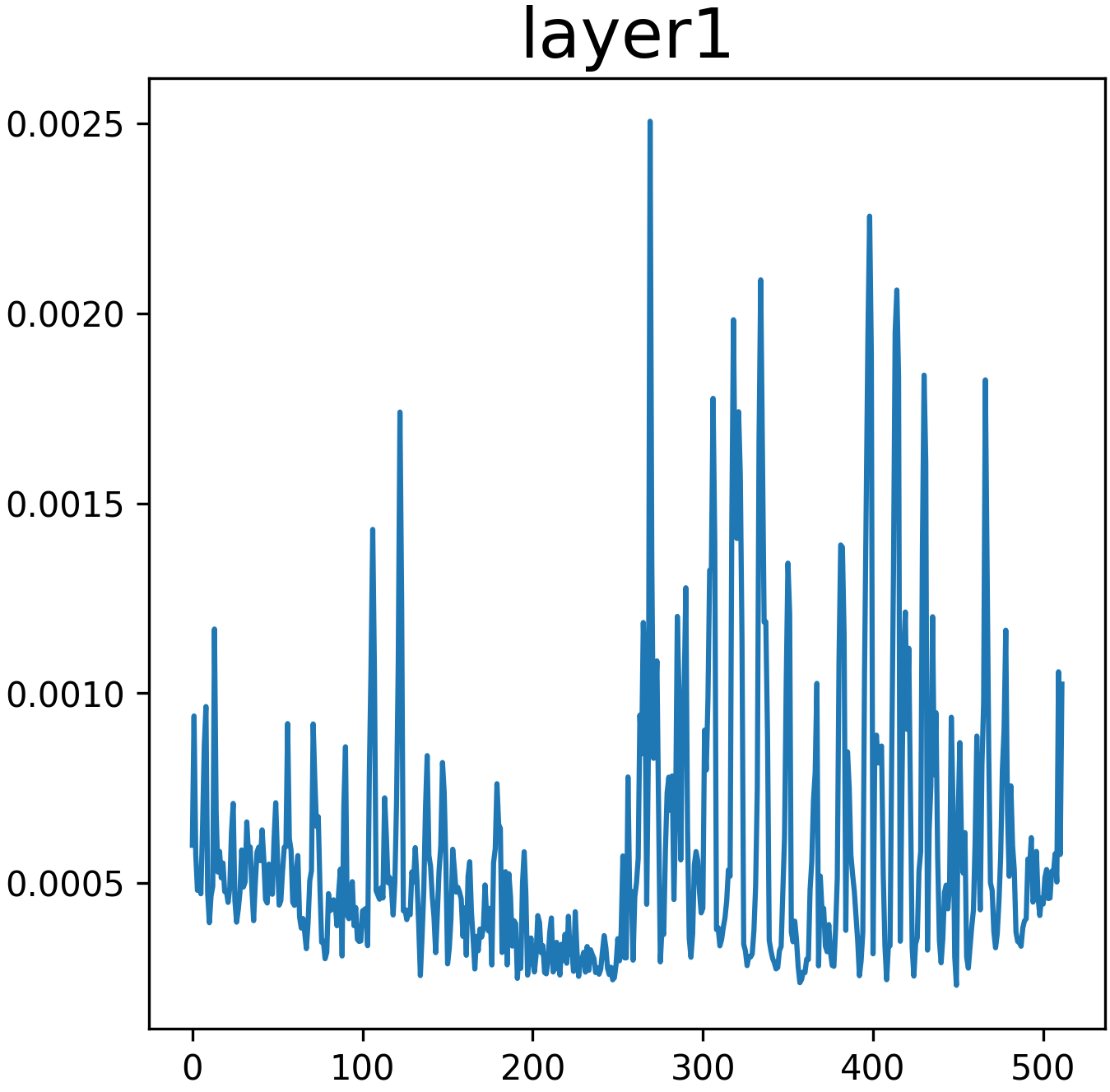}\\
        \includegraphics[width=0.405\textwidth]{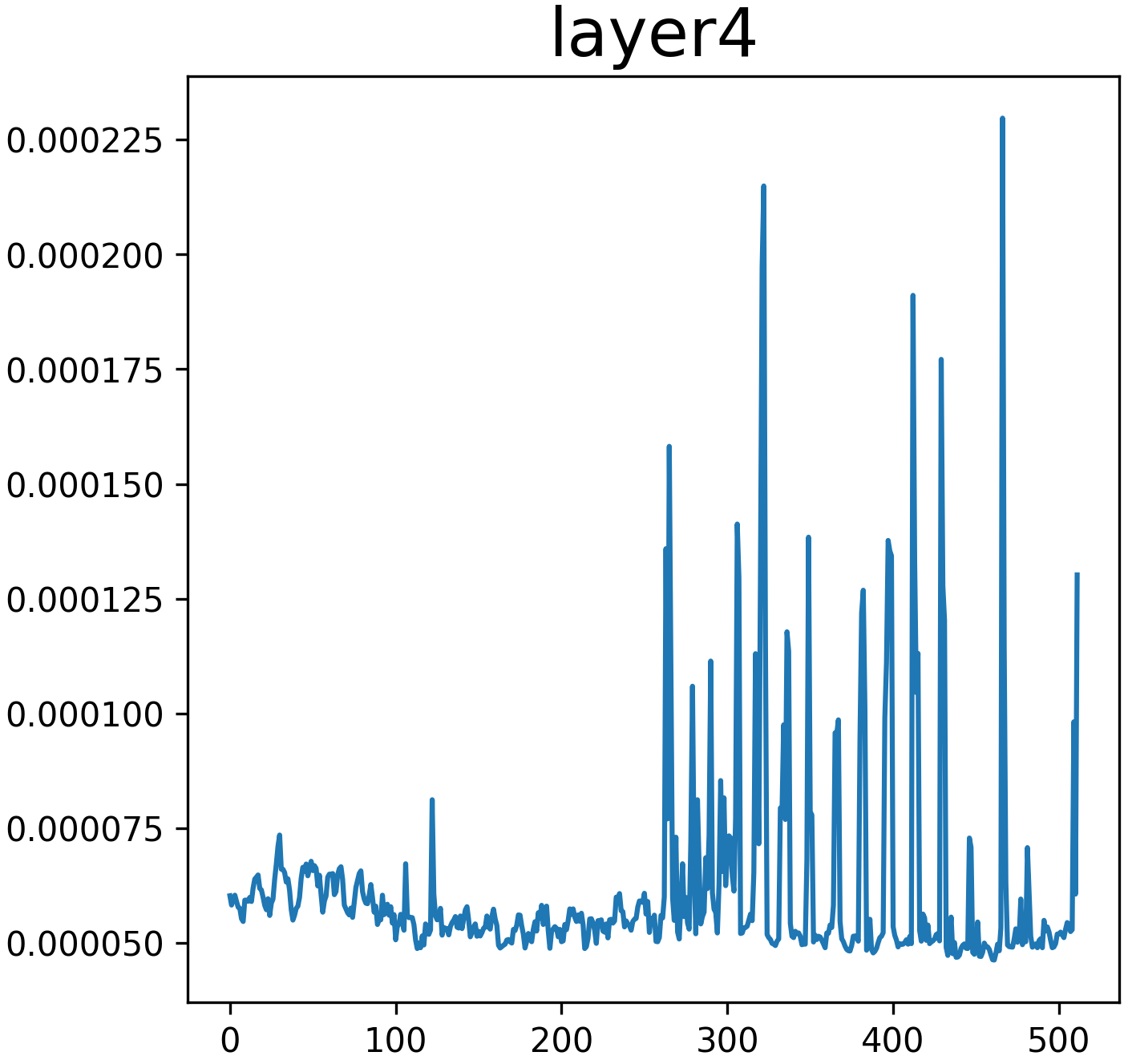}
        \includegraphics[width=0.405\textwidth]{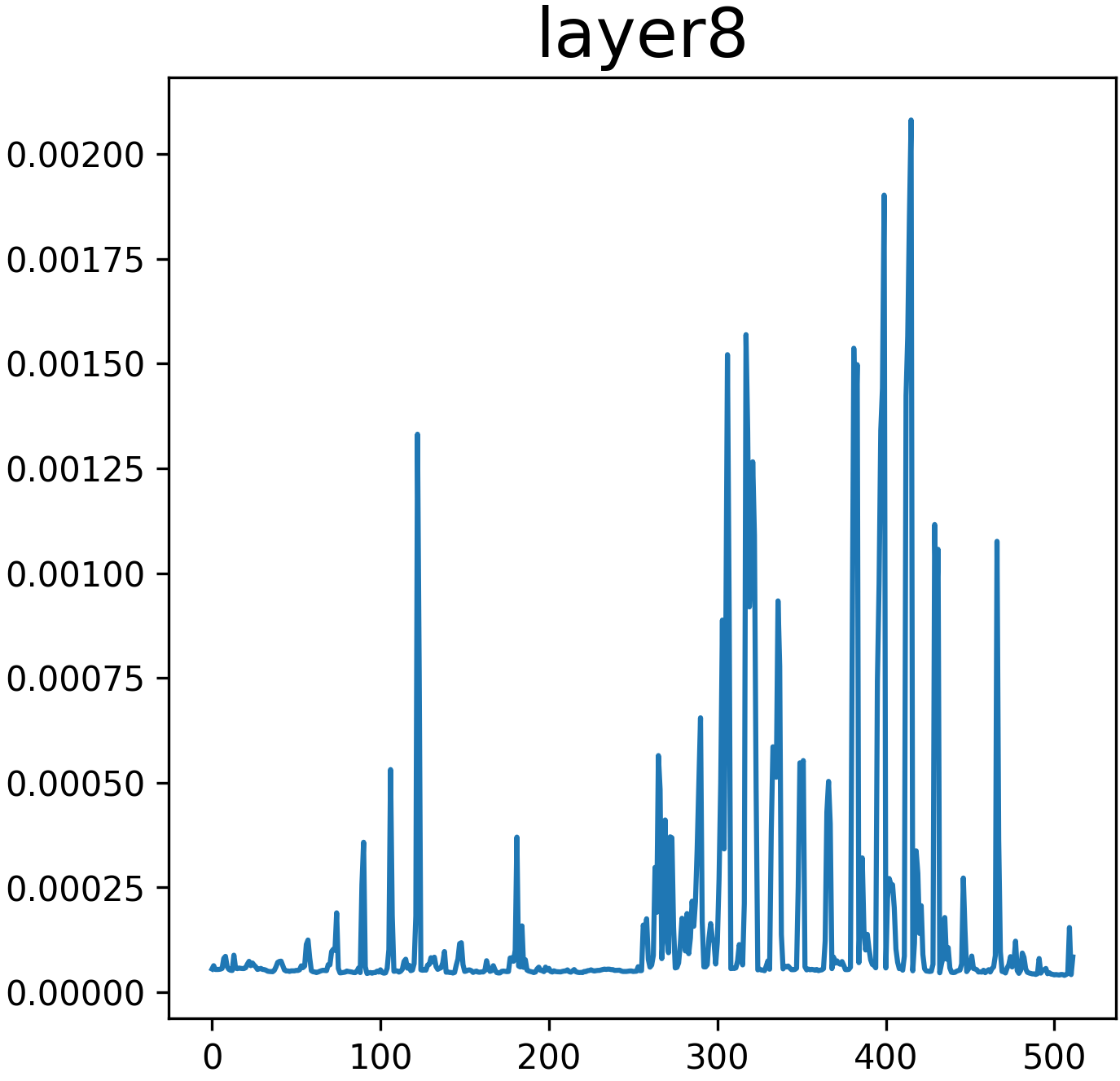}
        \caption{Vision tokens importance score. 
        % $\Phi$
        }
        \label{fig:vis_imp}
    \end{subfigure}
    \vspace{-0.1in}
    \caption{
    The figure illustrates the layer-wise importance score of the vision tokens. 
    (a) Vision–vision self-similarity (across layers)
    (b) Text→vision attention (after ADP)
    (c) Vision-token importance $\Phi$
    }
    \label{fig:scoremap}
    \vspace{-0.2in}
\end{figure*}
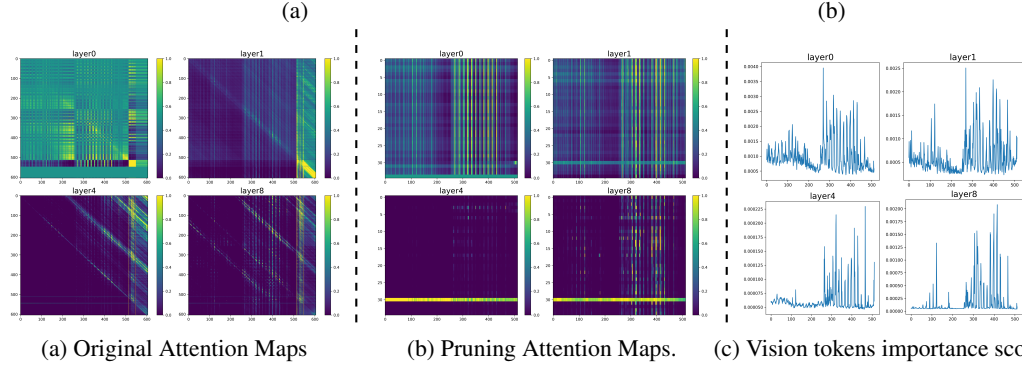

\textbf{Action-aware Dynamic Strategy.}
We assess the impact of the dynamic strategy with two baselines: removing the dynamic function (\emph{ADP w/o D}) and a handcrafted periodic switching variant (\emph{w/o D + PS}).
Table~\ref{tab:ablation_dynamic} (a) shows that introducing the dynamic controller (ADP) yields the best overall accuracy–compute balance: 96.3\% average SR with $\rho^{avg}=0.22$ and 6.43 FLOPs, outperforming the variant \emph{w/o D} (93.45\%, $\rho^{avg}=0.25$, 6.23 FLOPs) and the periodic schedule \emph{w/o D + PS} (89.9\%, $\rho^{avg}=0.50$, 4.55 FLOPs). Per-suite, the dynamic policy preserves Goal/Long while boosting Spatial/Object—most notably a \textbf{+16.6} point gain on \emph{Object} over the periodic schedule (98.0\% vs.\ 81.4\%), and \textbf{+4.4} on \emph{Spatial} (99.4\% vs.\ 95.0\%). Compared to \emph{ADP w/o D}, ADP improves average SR by \textbf{+2.85} points with comparable compute (+0.20 FLOPs), indicating that state-aware switching (rather than fixed cycling) is crucial to avoid over-pruning during fine manipulation while still pruning aggressively when large motions occur.

\textbf{Impact of Pruning Strategy.}
We ablate layer selection to validate using layer 0 for importance scoring. Table~\ref{tab:ablation_layer_rho} (b) shows that final SR is similar across layers, with layer 0 yielding the best accuracy–compute balance: 96.3\% average SR with 6.43 FLOPs, compared to layer 1 ($95.5\%$, $6.57$ FLOPs) and layer 4 ($95.8\%$, $6.89$ FLOPs). Meanwhile, FLOPs increase with deeper layers, so selecting $l=0$ offers a superior performance–efficiency trade-off. The slight decline in average SR with depth corroborates our analysis: deeper layers induce more localised attention—sharpening peaks vs.\ non-peaks but reducing global state coverage—thus becoming more sensitive to occasional mismatches and noise, which leads to marginally lower success rates.

\textbf{Observation of Attention Weights.}
A common view holds that text–vision alignment concentrates in deeper multimodal layers, so high importance scores should be computed there; however, this need not hold for parallel-decoding VLA. Our visualizations and measurements show that layer 0 already provides a stable, discriminative text-to-vision signal. In Fig.~\ref{fig:smss}, the layer-0 self-similarity matrix shows a clear high-contrast block structure, while deeper layers become increasingly diagonal-banded, compressing non-local correlations. The text to vision submatrix (Fig.~\ref{fig:smtv}) follows the same trend: layer 0 exhibits pronounced peaks and troughs across many vision tokens, whereas deeper layers retain only a few narrow high-response bands. The token importance score $\Phi$ (Fig.~\ref{fig:vis_imp}) likewise has higher SNR at layer 0; with depth, curves sharpen and develop long tails or near collapse, making top-$k$ ranking more sensitive to local noise.

%% file: sec/5_conclusion.tex
\section{Conclusion}
\label{sec:Conclusion}
We presented VLA-ADP, a plug-and-play pruning framework that unifies text-driven anticipatory pruning with an action-aware dynamic strategy to accelerate VLA inference while preserving reliability. Across simulation and real-world evaluations, the method maintains or improves task success, reduces compute, and shortens inference latency. The dynamic controller consistently outperforms removal and periodic switching, and early-layer scoring offers the best accuracy–efficiency balance. These results indicate that adaptively pruning vision tokens by motion state and instruction relevance enables efficient, fine-grained manipulation without compromising control quality.

%% file: sec/x_supp.tex
\section*{Appendix}
\newlength{\subfigH}
\setlength{\subfigH}{0.4\columnwidth} % 统一高度（可调）

\subsection*{Algorithm}
Here we provide the full pipeline of our proposed action-aware dynamic pruning for vision-language-action models.
\begin{algorithm}[htbp]
\caption{Action-Aware Dynamic Pruning (ADP) for VLA Inference}
\label{alg:adp}
\textbf{Require:} Multimodal embeddings $\mathbf{X}^m{=}\big[\mathbf{X}^{\text{[BOS]}},\mathbf{X}^{\text{vis}},\mathbf{X}^{\text{prop}},\mathbf{X}^{\text{txt}},\mathbf{X}^{\text{act}},\mathbf{X}^{\text{[EOS]}}\big]$,\\
\hspace*{3.4em} Q/K weights $W_Q^{(0)},W_K^{(0)}$, retention ratio $\rho$, view weights $\boldsymbol{\alpha}{\in}\mathbb{R}^C$ with $\sum_c \alpha_c{=}1$,\\
\hspace*{3.4em} gating rule $f(\cdot)$ (mean or adjacent-extrema), windowed actions $\mathbf{A}_i^{\,c}$, pose $T_{b_i}$\\
\textbf{Output:} Per-window visual state $s_{i}\!\in\!\{0,1\}$, pruned sequence $\tilde{\mathbf{X}}^{m}$, actions $\hat{\mathbf{a}}$
\begin{algorithmic}[1]
\State \textbf{Init:} $s_{1}\!\leftarrow\!0$ (cold start, full vision), $s_{2}\!\leftarrow\!0$; $\texttt{consec}\!\leftarrow\!0$ \Comment{two-window cold start}
\For{window $i=1,2,\dots$}
    \State \textbf{Compute windowed motion} $\delta_i$ via FK on $\mathbf{A}_i^{\,c}$ (Def.~\ref{def:fk_window}):
    \[
      \delta_i \leftarrow \sum_{t=b_i}^{e_i-1}\|\mathbf{p}_{t+1}-\mathbf{p}_t\|_2 \quad \text{with } \mathbf{p}_{t}=\pi(T_t),\ T_{t+1}=T_t
      \begin{bmatrix}R_{i,u} & \mathbf{v}_{i,u}\\ \mathbf{0}^\top & 1\end{bmatrix}
    \]
    \State \textbf{Gate update:} $s_{i+1}\leftarrow f(\delta_i)$ 
    \If{$\texttt{consec}\ge 3$} \Comment{reset to avoid prolonged pruning}
        \State $s_{i+1}\leftarrow 0$;\; $\texttt{consec}\leftarrow 0$
    \EndIf
    \State \textbf{Branch by state}
    \If{$s_{i+1}=1$} \Comment{pruned path}
        \State \textbf{Q/K scoring at layer 0:}
        \Statex\hspace{2.6em}$\mathbf{Q}^{(0)}\!=\!\mathbf{H}^{(0)}_{\text{txt}}W_Q^{(0)}$, $\mathbf{K}^{(0)}\!=\!\mathbf{H}^{(0)}_{\text{vis}}W_K^{(0)}$, 
        $\ \mathbf{A}^{(0)}\!=\!\frac{\mathbf{Q}^{(0)}(\mathbf{K}^{(0)})^\top}{\sqrt{d}}$
        \State \textbf{Aggregate importance} (Eq.~\ref{eq:attention score}): $\Phi^{(0)}(v)\leftarrow \frac{1}{N^h L_{\text{txt}}}\sum_{h,t}\mathbf{A}^{(0)}_{h,t,v}$
        \State $k\leftarrow \lfloor \rho\cdot L^{\text{vis}}\rfloor$;\; $k_c\leftarrow \lfloor \alpha_c\cdot k\rfloor$ for $c{=}1..C$
        \State \textbf{Per-view Top-$k$:} $\mathbf{X}^{\text{vis}}_{(c)}\leftarrow \operatorname{Top\mbox{-}K}(\Phi^{(0)}_{(c)},k_c)$;\;
               $\mathbf{X}^{\text{vis}}_{\text{keep}}\leftarrow \bigcup_{c}\mathbf{X}^{\text{vis}}_{(c)}$
        \State \textbf{Form pruned input:} 
               $\tilde{\mathbf{X}}^{m}\!\leftarrow\![\mathbf{X}^{\text{[BOS]}},\mathbf{X}^{\text{vis}}_{\text{keep}},\mathbf{X}^{\text{prop}},\mathbf{X}^{\text{txt}},\mathbf{X}^{\text{act}},\mathbf{X}^{\text{[EOS]}}]$
        \State \textbf{LLM forward:} $\hat{\mathbf{a}}\leftarrow f_{\text{LLM}}(\tilde{\mathbf{X}}^{m})$;\; $\texttt{consec}\leftarrow \texttt{consec}+1$
    \Else \Comment{full-vision path}
        \State $\tilde{\mathbf{X}}^{m}\leftarrow \mathbf{X}^{m}$;\; $\hat{\mathbf{a}}\leftarrow f_{\text{LLM}}(\mathbf{X}^{m})$;\; $\texttt{consec}\leftarrow 0$
    \EndIf
    \State \textbf{Execute} $\hat{\mathbf{a}}$ and advance to next window
\EndFor
\State \Return $\{s_i\}$, $\tilde{\mathbf{X}}^{m}$, $\hat{\mathbf{a}}$
\end{algorithmic}
\end{algorithm}
\vspace{-0.1in}

\subsection*{Real-World Experiment Details}

We provide additional details of the real-robot experiments to facilitate reproducibility.

\paragraph{Robot Platform.}
All real-world evaluations were conducted on a \textbf{Kinova Jaco2} 6-DoF robotic arm equipped with a parallel-jaw gripper. The robot was controlled through Cartesian velocity commands at 10\,Hz, with joint and workspace safety limits enforced to prevent collisions.

\paragraph{Sensing Setup.}
A single RGB camera (Sony AX53) was mounted in front of the table to capture the entire workspace. The camera streamed RGB frames at $640{\times}480$ resolution and 30\,FPS. No wrist-mounted camera was used in the real-robot experiments, and all observations were obtained from this fixed view.

\paragraph{Runtime Environment.}
All policies were executed on a Linux workstation running Ubuntu 20.04 with an NVIDIA RTX~4090 GPU. Inference employed parallel decoding with an OFT chunk size of 8. Latency was measured as mean per-step inference time over 100 runs.

\paragraph{Task Setup.}
We evaluated four tabletop manipulation tasks:
\textbf{a) Task 1:} Pick up the orange pot and place it into a green plate.
\textbf{b) Task 2:} Pick and place a blue cube onto the plate.
\textbf{c) Task 3:} Pick and place a carrot into a blue pan.
\textbf{d) Task 4:} Wipe the table.
Objects were placed randomly within a $50{\times}50$\,cm workspace area at the beginning of each trial. 
For data collection, each task was executed using a gamepad teleoperation interface, with the human operator directly controlling the Jaco2 arm. We collected approximately 100--150 trajectories per task, covering diverse initial states and object positions. 
The collected demonstrations were then used to build the training dataset, and we fine-tuned OpenVLA-OFT separately on each task before deploying the model for evaluation in the real-robot experiments.

\paragraph{Evaluation Protocol.}
Each task was repeated 30 trials under randomized object initializations. A trial was marked as successful if the object was placed entirely within the target receptacle (for Tasks~1--3) or if more than $80\%$ of the designated area was wiped (Task~4). At the end of each episode, the robot was reset to a neutral home pose. Safety termination was triggered if joint torque exceeded preset thresholds.

\subsection*{Quantitative Statistics}

To complement the attention visualizations with comparable quantitative evidence, we compute scale-free statistics at each layer from the text-to-vision importance vector $\Phi=\{\phi_i\}_{i=1}^{V}$. We first normalize $\Phi$ as $p_i=\phi_i/\sum_{j=1}^{V}\phi_j$ and then evaluate the \emph{Participation Ratio (PR)} and the \emph{Entropy ($\mathcal{H}$)}:
\begin{equation}
    PR=\frac{1}{\sum_{i=1}^{V}p_i^2},\quad
    \mathcal{H}(p)=-\sum_{i=1}^{V}p_i\log p_i .
\end{equation}
$PR$ approximates the effective number of participating visual columns, while $\mathcal{H}$ quantifies the dispersion and multi-modality of the distribution. Together, they characterize the spatial coverage of alignment signals and the stability of Top-$K$ ranking.
 
\begin{figure}[th]
  \centering
  \vspace{-0.1in}
  \resizebox{0.8\textwidth}{!}{
  \begin{subfigure}[t]{0.45\columnwidth}
    \centering    \includegraphics[height=\subfigH,keepaspectratio]{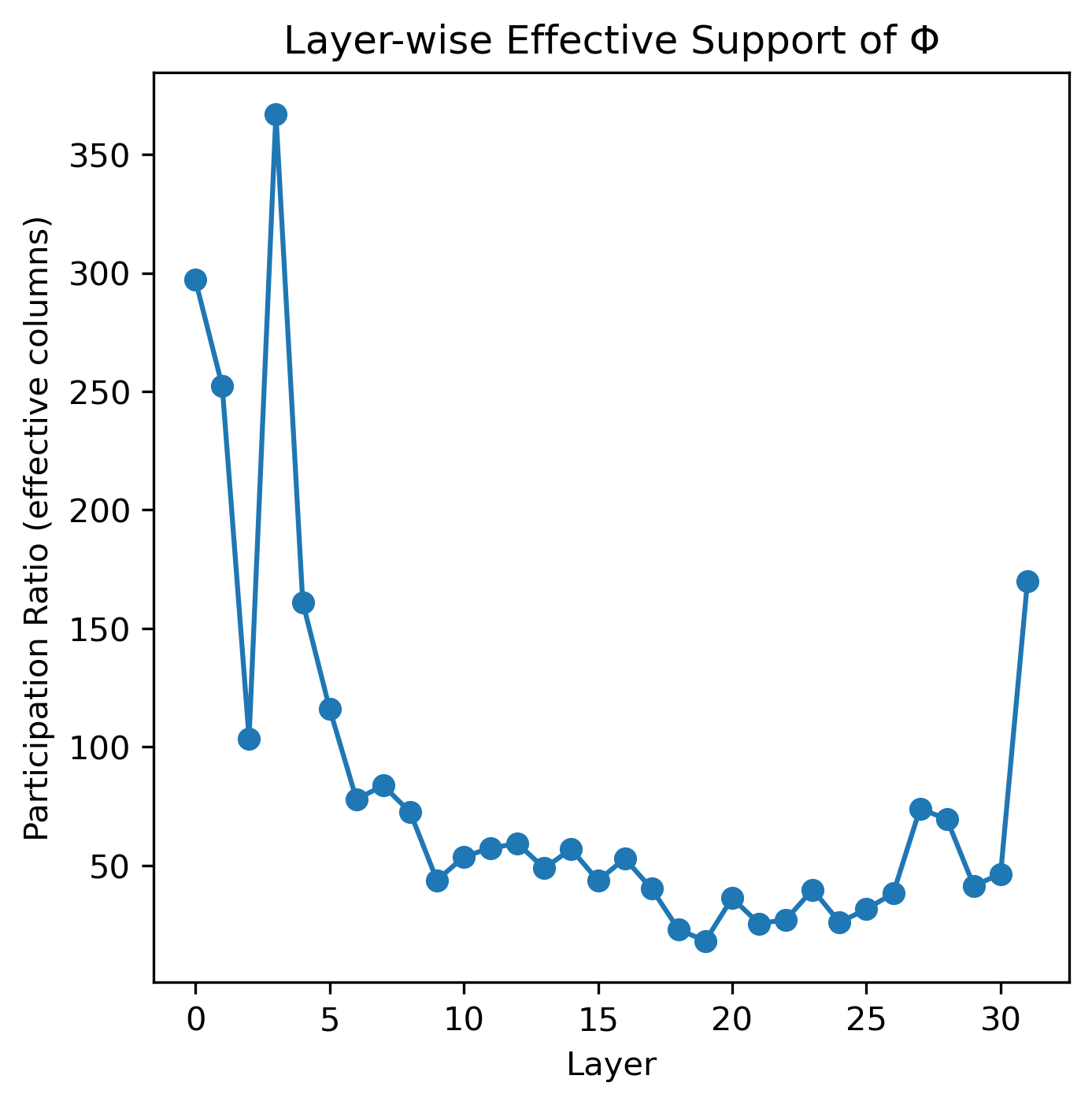}
    \caption{$PR$ across layers}
    \label{fig:layer_pr}
  \end{subfigure}
  \hfill
  \begin{subfigure}[t]{0.45\columnwidth}
    \centering
\includegraphics[height=\subfigH,keepaspectratio]{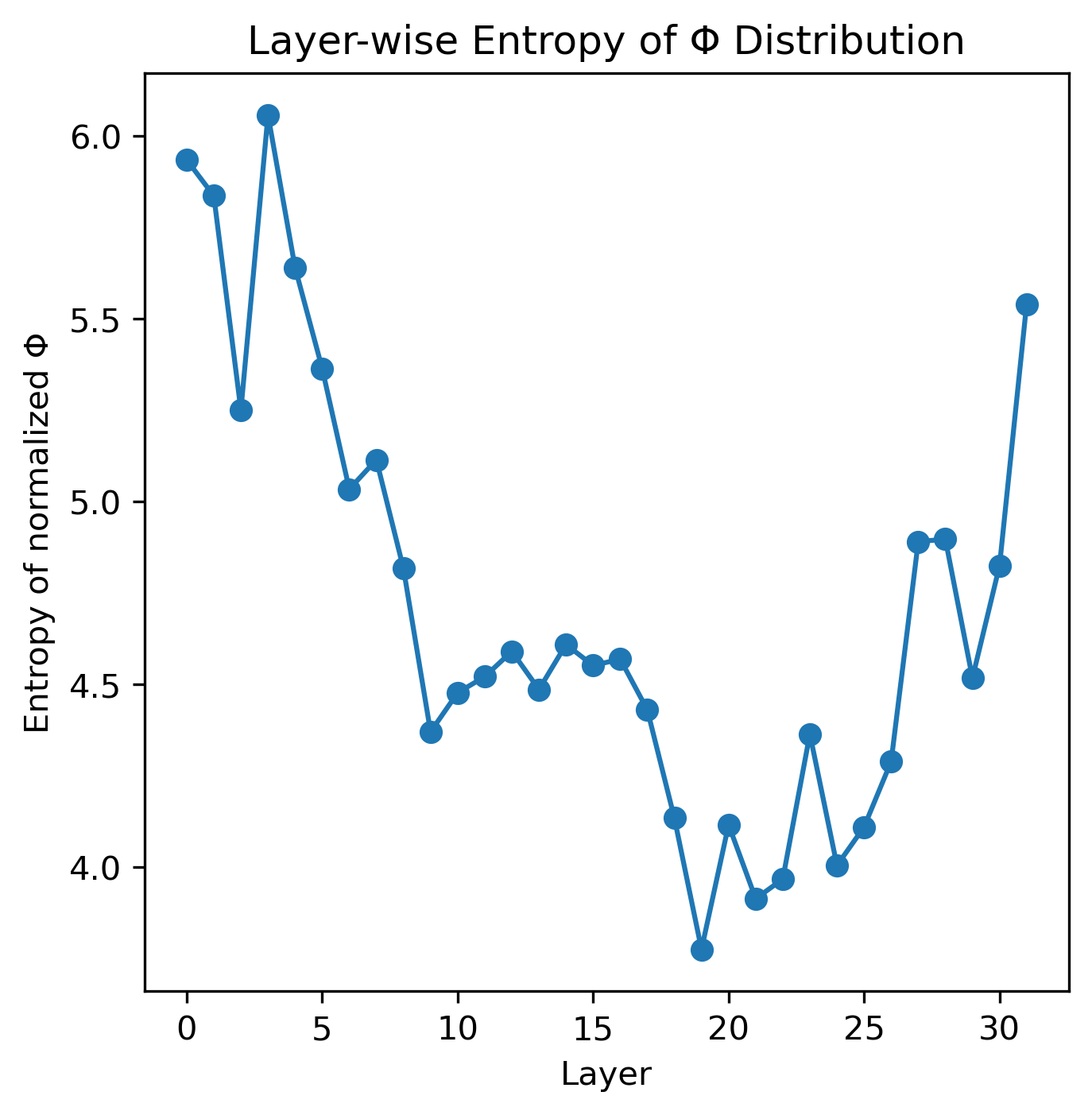}
    \caption{Entropy $\mathcal{H}$ across layers}
    \label{fig:layer_entropy}
  \end{subfigure}
  }
  \vspace{-0.01in}
  \caption{Layer-wise statistics of text-to-vision importance scores.}
  \label{fig:layerstats_ab}
  \vspace{-0.1in}
\end{figure}

Figure \ref{fig:layer_pr} presents the variation of \textit{PR} across layers. In shallow layers, \textit{PR} remains relatively high, indicating that a large number of key vision tokens receive balanced attention and contribute to broad spatial coverage. As depth increases, \textit{PR} rapidly decreases and stabilizes, reflecting a progressive collapse of attention onto a small subset of tokens and reduced feature utilization.

Figure \ref{fig:layer_entropy} presents the corresponding entropy $\mathcal{H}$ across layers. Shallow layers exhibit higher entropy, suggesting diverse and smooth distributions that make Top-$K$ selection robust against individual outliers. In contrast, deeper layers display sharply peaked distributions dominated by a few tokens. While such localization may benefit semantic alignment in downstream reasoning, it renders Top-$K$ selection unstable and prone to discarding critical tokens.
These findings suggest that shallow-layer attention signals are preferable for estimating the importance of vision tokens in VLA tasks. In particular, using layer-0 offers a computationally efficient choice, providing reliable importance estimates while further reducing FLOPs.

\begin{figure}[th]
  \centering
  \resizebox{0.6\textwidth}{!}{
  \includegraphics[height=\subfigH,keepaspectratio]{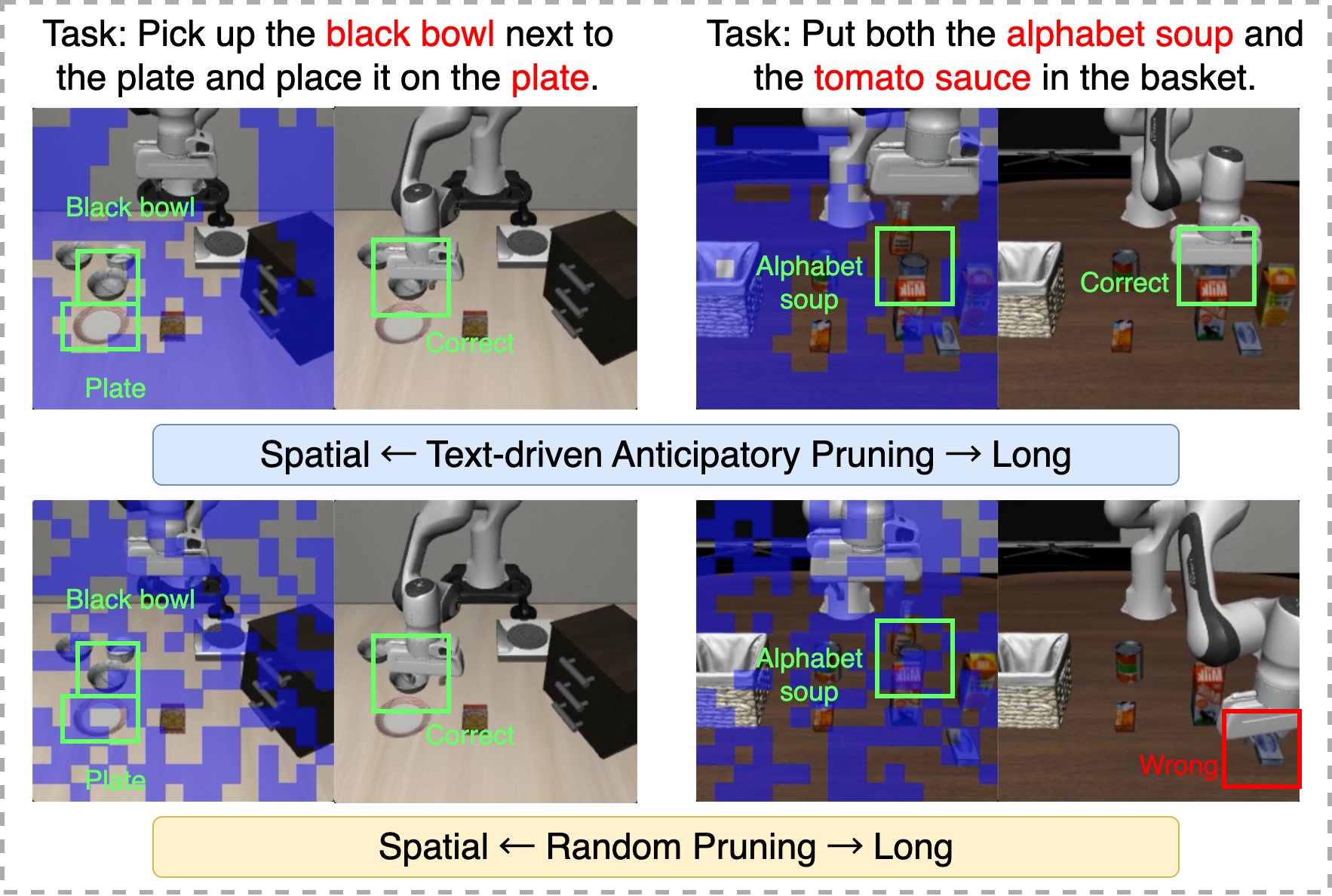}
  }
  \caption{ADP vs. Random Pruning.}
  \label{fig:random}
\end{figure}

\noindent\textbf{Compare with Random Pruning.} Random Pruning yields competitive performance on Spatial and Goal, but performs substantially worse on Object and Long. We can attribute this phenomenon to the distribution of visual patches spanned by the key object as shown in Figure \ref{fig:random}. When the target object spans a large number of patches, random pruning still has a high probability of retaining the relevant patches, thereby preserving sufficient task-critical vision information. In contrast, random pruning is more likely to prune the relevant patches when the key object spans only a small number of patches. Such pruning may result in incorrect object identification, which in turn propagates errors and ultimately causes task failure.

%% file: iclr2026_conference.bbl
\begin{thebibliography}{31}
\providecommand{\natexlab}[1]{#1}
\providecommand{\url}[1]{\texttt{#1}}
\expandafter\ifx\csname urlstyle\endcsname\relax
  \providecommand{\doi}[1]{doi: #1}\else
  \providecommand{\doi}{doi: \begingroup \urlstyle{rm}\Url}\fi

\bibitem[Awadalla et~al.(2023)Awadalla, Gao, Gardner, Hessel, Hanafy, Zhu, Marathe, Bitton, Gadre, Sagawa, et~al.]{awadalla2023openflamingo}
Anas Awadalla, Irena Gao, Josh Gardner, Jack Hessel, Yusuf Hanafy, Wanrong Zhu, Kalyani Marathe, Yonatan Bitton, Samir Gadre, Shiori Sagawa, et~al.
\newblock Openflamingo: An open-source framework for training large autoregressive vision-language models.
\newblock \emph{arXiv preprint arXiv:2308.01390}, 2023.

\bibitem[Bjorck et~al.(2025)Bjorck, Casta{\~n}eda, Cherniadev, Da, Ding, Fan, Fang, Fox, Hu, Huang, et~al.]{bjorck2025gr00t}
Johan Bjorck, Fernando Casta{\~n}eda, Nikita Cherniadev, Xingye Da, Runyu Ding, Linxi Fan, Yu~Fang, Dieter Fox, Fengyuan Hu, Spencer Huang, et~al.
\newblock Gr00t n1: An open foundation model for generalist humanoid robots.
\newblock \emph{arXiv preprint arXiv:2503.14734}, 2025.

\bibitem[Black et~al.(2024)Black, Brown, Driess, Esmail, Equi, Finn, Fusai, Groom, Hausman, Ichter, et~al.]{black2024pi}
Kevin Black, Noah Brown, Danny Driess, Adnan Esmail, Michael Equi, Chelsea Finn, Niccolo Fusai, Lachy Groom, Karol Hausman, Brian Ichter, et~al.
\newblock A vision-language-action flow model for general robot control.
\newblock \emph{arXiv preprint arXiv:2410.24164}, 2024.

\bibitem[Brohan et~al.(2024)Brohan, Brown, Carbajal, Chebotar, Chen, Choromanski, Ding, Driess, Dubey, Finn, et~al.]{brohan2024rt}
Anthony Brohan, Noah Brown, Justice Carbajal, Yevgen Chebotar, Xi~Chen, Krzysztof Choromanski, Tianli Ding, Danny Driess, Avinava Dubey, Chelsea Finn, et~al.
\newblock Rt-2: Vision-language-action models transfer web knowledge to robotic control, 2023.
\newblock \emph{URL https://arxiv. org/abs/2307.15818}, 2024.

\bibitem[Cen et~al.(2025)Cen, Yu, Yuan, Jiang, Huang, Guo, Li, Song, Luo, Wang, et~al.]{cen2025worldvla}
Jun Cen, Chaohui Yu, Hangjie Yuan, Yuming Jiang, Siteng Huang, Jiayan Guo, Xin Li, Yibing Song, Hao Luo, Fan Wang, et~al.
\newblock Worldvla: Towards autoregressive action world model.
\newblock \emph{arXiv preprint arXiv:2506.21539}, 2025.

\bibitem[Chen et~al.(2024)Chen, Zhao, Liu, Bai, Lin, Zhou, and Chang]{chen2024image}
Liang Chen, Haozhe Zhao, Tianyu Liu, Shuai Bai, Junyang Lin, Chang Zhou, and Baobao Chang.
\newblock An image is worth 1/2 tokens after layer 2: Plug-and-play inference acceleration for large vision-language models.
\newblock In \emph{European Conference on Computer Vision}, pp.\  19--35. Springer, 2024.

\bibitem[Chi et~al.(2023)Chi, Xu, Feng, Cousineau, Du, Burchfiel, Tedrake, and Song]{chi2023diffusion}
Cheng Chi, Zhenjia Xu, Siyuan Feng, Eric Cousineau, Yilun Du, Benjamin Burchfiel, Russ Tedrake, and Shuran Song.
\newblock Diffusion policy: Visuomotor policy learning via action diffusion.
\newblock \emph{The International Journal of Robotics Research}, pp.\  02783649241273668, 2023.

\bibitem[Hung et~al.(2025)Hung, Sun, Hong, Zadeh, Li, Tan, Majumder, Poria, et~al.]{hung2025nora}
Chia-Yu Hung, Qi~Sun, Pengfei Hong, Amir Zadeh, Chuan Li, U~Tan, Navonil Majumder, Soujanya Poria, et~al.
\newblock Nora: A small open-sourced generalist vision language action model for embodied tasks.
\newblock \emph{arXiv preprint arXiv:2504.19854}, 2025.

\bibitem[Kim et~al.(2024)Kim, Pertsch, Karamcheti, Xiao, Balakrishna, Nair, Rafailov, Foster, Lam, Sanketi, et~al.]{kim2024openvla}
Moo~Jin Kim, Karl Pertsch, Siddharth Karamcheti, Ted Xiao, Ashwin Balakrishna, Suraj Nair, Rafael Rafailov, Ethan Foster, Grace Lam, Pannag Sanketi, et~al.
\newblock Openvla: An open-source vision-language-action model.
\newblock \emph{arXiv preprint arXiv:2406.09246}, 2024.

\bibitem[Kim et~al.(2025)Kim, Finn, and Liang]{kim2025fine}
Moo~Jin Kim, Chelsea Finn, and Percy Liang.
\newblock Fine-tuning vision-language-action models: Optimizing speed and success.
\newblock \emph{arXiv preprint arXiv:2502.19645}, 2025.

\bibitem[Li et~al.(2024)Li, Liang, Wang, Luo, Chen, Liao, Wei, Deng, Xu, Zhang, et~al.]{li2024cogact}
Qixiu Li, Yaobo Liang, Zeyu Wang, Lin Luo, Xi~Chen, Mozheng Liao, Fangyun Wei, Yu~Deng, Sicheng Xu, Yizhong Zhang, et~al.
\newblock Cogact: A foundational vision-language-action model for synergizing cognition and action in robotic manipulation.
\newblock \emph{arXiv preprint arXiv:2411.19650}, 2024.

\bibitem[Li et~al.(2025)Li, Meng, Sun, Ji, Tang, Fan, Ma, Xia, Wang, and Zhu]{li2025sp}
Ye~Li, Yuan Meng, Zewen Sun, Kangye Ji, Chen Tang, Jiajun Fan, Xinzhu Ma, Shutao Xia, Zhi Wang, and Wenwu Zhu.
\newblock Sp-vla: A joint model scheduling and token pruning approach for vla model acceleration.
\newblock \emph{arXiv preprint arXiv:2506.12723}, 2025.

\bibitem[Liu et~al.(2023{\natexlab{a}})Liu, Zhu, Gao, Feng, Liu, Zhu, and Stone]{liu2023libero}
Bo~Liu, Yifeng Zhu, Chongkai Gao, Yihao Feng, Qiang Liu, Yuke Zhu, and Peter Stone.
\newblock Libero: Benchmarking knowledge transfer for lifelong robot learning.
\newblock \emph{Advances in Neural Information Processing Systems}, 36:\penalty0 44776--44791, 2023{\natexlab{a}}.

\bibitem[Liu et~al.(2023{\natexlab{b}})Liu, Li, Li, and Lee]{liu2023improvedllava}
Haotian Liu, Chunyuan Li, Yuheng Li, and Yong~Jae Lee.
\newblock Improved baselines with visual instruction tuning, 2023{\natexlab{b}}.

\bibitem[Liu et~al.(2023{\natexlab{c}})Liu, Li, Wu, and Lee]{liu2023llava}
Haotian Liu, Chunyuan Li, Qingyang Wu, and Yong~Jae Lee.
\newblock Visual instruction tuning, 2023{\natexlab{c}}.

\bibitem[Liu et~al.(2024{\natexlab{a}})Liu, Li, Li, Li, Zhang, Shen, and Lee]{liu2024llavanext}
Haotian Liu, Chunyuan Li, Yuheng Li, Bo~Li, Yuanhan Zhang, Sheng Shen, and Yong~Jae Lee.
\newblock Llava-next: Improved reasoning, ocr, and world knowledge, January 2024{\natexlab{a}}.
\newblock URL \url{https://llava-vl.github.io/blog/2024-01-30-llava-next/}.

\bibitem[Liu et~al.(2024{\natexlab{b}})Liu, Liu, Wang, Lee, Zhou, An, Yang, Zhang, Guo, and Zhang]{liu2024robomamba}
Jiaming Liu, Mengzhen Liu, Zhenyu Wang, Lily Lee, Kaichen Zhou, Pengju An, Senqiao Yang, Renrui Zhang, Yandong Guo, and Shanghang Zhang.
\newblock Robomamba: Multimodal state space model for efficient robot reasoning and manipulation.
\newblock \emph{arXiv e-prints}, pp.\  arXiv--2406, 2024{\natexlab{b}}.

\bibitem[Oquab et~al.(2023)Oquab, Darcet, Moutakanni, Vo, Szafraniec, Khalidov, Fernandez, Haziza, Massa, El-Nouby, et~al.]{oquab2023dinov2}
Maxime Oquab, Timoth{\'e}e Darcet, Th{\'e}o Moutakanni, Huy Vo, Marc Szafraniec, Vasil Khalidov, Pierre Fernandez, Daniel Haziza, Francisco Massa, Alaaeldin El-Nouby, et~al.
\newblock Dinov2: Learning robust visual features without supervision.
\newblock \emph{arXiv preprint arXiv:2304.07193}, 2023.

\bibitem[Pei et~al.(2024)Pei, Huang, and Xu]{pei2024cross}
Xiaohuan Pei, Tao Huang, and Chang Xu.
\newblock Cross-self kv cache pruning for efficient vision-language inference.
\newblock \emph{arXiv preprint arXiv:2412.04652}, 2024.

\bibitem[Shukor et~al.(2025)Shukor, Aubakirova, Capuano, Kooijmans, Palma, Zouitine, Aractingi, Pascal, Russi, Marafioti, et~al.]{shukor2025smolvla}
Mustafa Shukor, Dana Aubakirova, Francesco Capuano, Pepijn Kooijmans, Steven Palma, Adil Zouitine, Michel Aractingi, Caroline Pascal, Martino Russi, Andres Marafioti, et~al.
\newblock Smolvla: A vision-language-action model for affordable and efficient robotics.
\newblock \emph{arXiv preprint arXiv:2506.01844}, 2025.

\bibitem[Tan et~al.(2025)Tan, Yang, Ye, Zheng, Bai, Wang, Hao, and Chen]{tan2025think}
Xudong Tan, Yaoxin Yang, Peng Ye, Jialin Zheng, Bizhe Bai, Xinyi Wang, Jia Hao, and Tao Chen.
\newblock Think twice, act once: Token-aware compression and action reuse for efficient inference in vision-language-action models.
\newblock \emph{arXiv preprint arXiv:2505.21200}, 2025.

\bibitem[Team et~al.(2023)Team, Anil, Borgeaud, Alayrac, Yu, Soricut, Schalkwyk, Dai, Hauth, Millican, et~al.]{team2023gemini}
Gemini Team, Rohan Anil, Sebastian Borgeaud, Jean-Baptiste Alayrac, Jiahui Yu, Radu Soricut, Johan Schalkwyk, Andrew~M Dai, Anja Hauth, Katie Millican, et~al.
\newblock Gemini: a family of highly capable multimodal models.
\newblock \emph{arXiv preprint arXiv:2312.11805}, 2023.

\bibitem[Touvron et~al.(2023)Touvron, Martin, Stone, Albert, Almahairi, Babaei, Bashlykov, Batra, Bhargava, Bhosale, et~al.]{touvron2023llama}
Hugo Touvron, Louis Martin, Kevin Stone, Peter Albert, Amjad Almahairi, Yasmine Babaei, Nikolay Bashlykov, Soumya Batra, Prajjwal Bhargava, Shruti Bhosale, et~al.
\newblock Llama 2: Open foundation and fine-tuned chat models.
\newblock \emph{arXiv preprint arXiv:2307.09288}, 2023.

\bibitem[Wen et~al.(2025{\natexlab{a}})Wen, Zhu, Li, Tang, Shen, and Feng]{wen2025dexvla}
Junjie Wen, Yichen Zhu, Jinming Li, Zhibin Tang, Chaomin Shen, and Feifei Feng.
\newblock Dexvla: Vision-language model with plug-in diffusion expert for general robot control.
\newblock \emph{arXiv preprint arXiv:2502.05855}, 2025{\natexlab{a}}.

\bibitem[Wen et~al.(2025{\natexlab{b}})Wen, Zhu, Zhu, Tang, Li, Zhou, Liu, Shen, Peng, and Feng]{wen2025diffusionvla}
Junjie Wen, Yichen Zhu, Minjie Zhu, Zhibin Tang, Jinming Li, Zhongyi Zhou, Xiaoyu Liu, Chaomin Shen, Yaxin Peng, and Feifei Feng.
\newblock Diffusionvla: Scaling robot foundation models via unified diffusion and autoregression.
\newblock In \emph{Forty-second International Conference on Machine Learning}, 2025{\natexlab{b}}.

\bibitem[Xu et~al.(2025)Xu, Wang, Xia, Zhu, Huang, and Xu]{xu2025vla}
Siyu Xu, Yunke Wang, Chenghao Xia, Dihao Zhu, Tao Huang, and Chang Xu.
\newblock Vla-cache: Towards efficient vision-language-action model via adaptive token caching in robotic manipulation.
\newblock \emph{arXiv preprint arXiv:2502.02175}, 2025.

\bibitem[Yang et~al.(2025)Yang, Wang, Wen, Zhongwei, Zou, Zhang, Wen, and Zhang]{yang2025efficientvla}
Yantai Yang, Yuhao Wang, Zichen Wen, Luo Zhongwei, Chang Zou, Zhipeng Zhang, Chuan Wen, and Linfeng Zhang.
\newblock Efficientvla: Training-free acceleration and compression for vision-language-action models.
\newblock \emph{arXiv preprint arXiv:2506.10100}, 2025.

\bibitem[Yue et~al.(2024)Yue, Wang, Kang, Han, Wang, Song, Feng, and Huang]{yue2024deer}
Yang Yue, Yulin Wang, Bingyi Kang, Yizeng Han, Shenzhi Wang, Shiji Song, Jiashi Feng, and Gao Huang.
\newblock Deer-vla: Dynamic inference of multimodal large language models for efficient robot execution.
\newblock \emph{Advances in Neural Information Processing Systems}, 37:\penalty0 56619--56643, 2024.

\bibitem[Zhai et~al.(2023)Zhai, Mustafa, Kolesnikov, and Beyer]{zhai2023sigmoid}
Xiaohua Zhai, Basil Mustafa, Alexander Kolesnikov, and Lucas Beyer.
\newblock Sigmoid loss for language image pre-training.
\newblock In \emph{Proceedings of the IEEE/CVF international conference on computer vision}, pp.\  11975--11986, 2023.

\bibitem[Zhang et~al.(2025)Zhang, Dong, Zhang, Heng, Chi, Dai, Du, Du, and Zhang]{zhang2025mole}
Rongyu Zhang, Menghang Dong, Yuan Zhang, Liang Heng, Xiaowei Chi, Gaole Dai, Li~Du, Yuan Du, and Shanghang Zhang.
\newblock Mole-vla: Dynamic layer-skipping vision language action model via mixture-of-layers for efficient robot manipulation.
\newblock \emph{arXiv preprint arXiv:2503.20384}, 2025.

\bibitem[Zhang et~al.(2024)Zhang, Fan, Ma, Zheng, Huang, Cheng, Gudovskiy, Okuno, Nakata, Keutzer, et~al.]{zhang2024sparsevlm}
Yuan Zhang, Chun-Kai Fan, Junpeng Ma, Wenzhao Zheng, Tao Huang, Kuan Cheng, Denis Gudovskiy, Tomoyuki Okuno, Yohei Nakata, Kurt Keutzer, et~al.
\newblock Sparsevlm: Visual token sparsification for efficient vision-language model inference.
\newblock \emph{arXiv preprint arXiv:2410.04417}, 2024.

\end{thebibliography}
